\documentclass{article}
\usepackage{graphicx}
\usepackage{subcaption}
\usepackage{caption}

\usepackage{PRIMEarxiv}

\usepackage[utf8]{inputenc} 
\usepackage[T1]{fontenc}    
\usepackage{hyperref}       
\usepackage{url}            
\usepackage{booktabs}       
\usepackage{amsfonts}
\usepackage{amsmath}
\usepackage{tabularx}
\usepackage{nicefrac}       
\usepackage{microtype}      
\usepackage{lipsum}
\usepackage{wrapfig}
\usepackage{array}
\usepackage[linesnumbered,ruled,vlined]{algorithm2e}
\usepackage{fancyhdr}  
\usepackage{authblk}
\usepackage{multirow}
\usepackage{graphicx}       
\graphicspath{{media/}}     
\usepackage{hyperref}
\usepackage{academicons}
\usepackage{xcolor} 
\usepackage{lipsum}
\newcommand{\orcidlink}[2]{\href{https://orcid.org/#2}{\textbf{#1} \hspace{1mm} \includegraphics[height=0.8em]{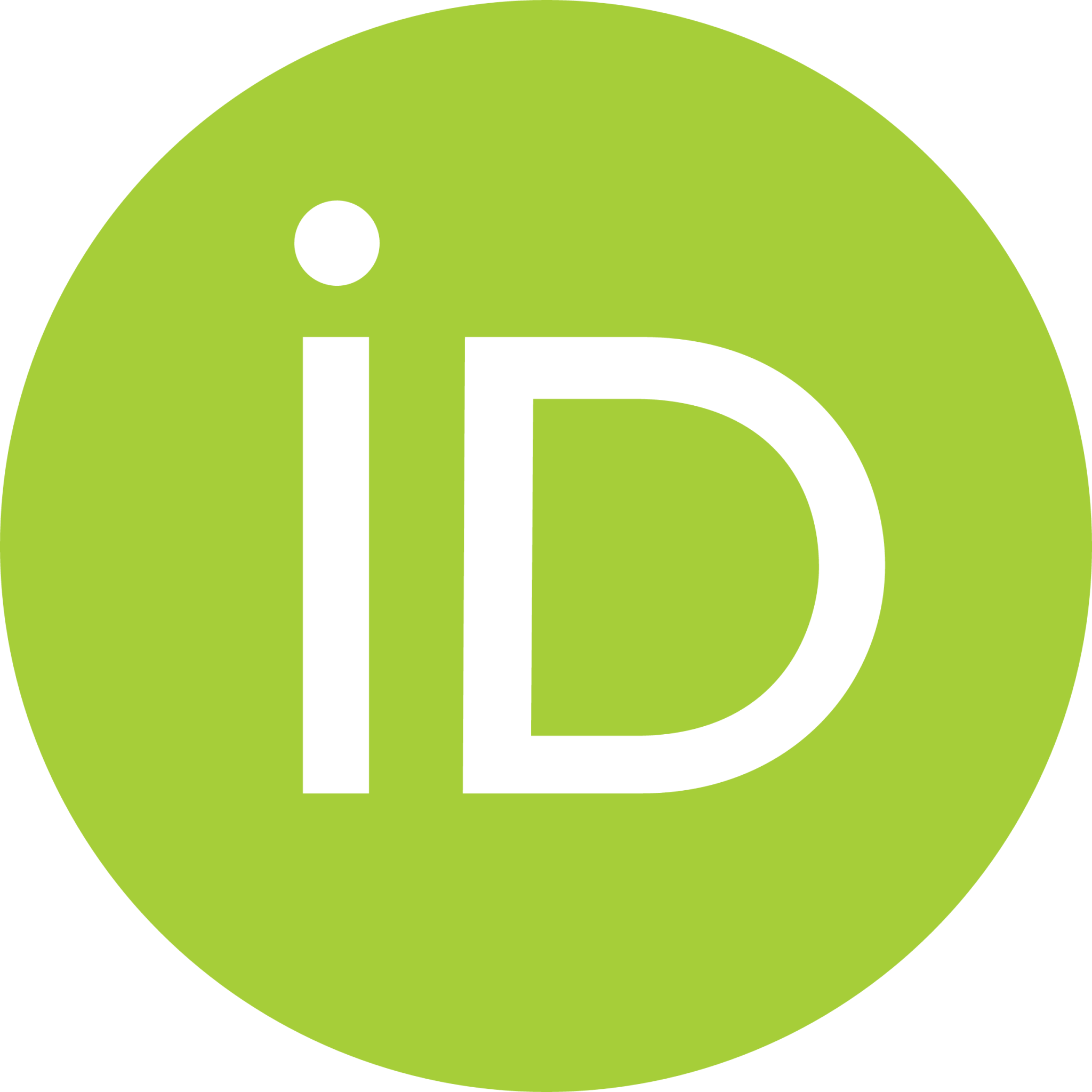}}}

\title{NutriVision: A System for Automatic Diet Management in Smart Healthcare}
\author{}
\date{}
\begin{document}
\maketitle

\vspace{-2cm}
\begin{center}
    \orcidlink{Madhumita Veeramreddy}{0009-0004-9279-449X} \\
    Department of Computer Science \& Engineering \\
    SRM University AP, Amaravati, India \\
    \href{mailto:madhumita_v@srmap.edu.in}{\texttt{madhumita\_v@srmap.edu.in}}
\end{center}

\vspace{0.15cm}

\begin{center}
\begin{tabular}{cc}
\begin{tabular}{c}
\orcidlink{Ashok Kumar Pradhan}{0000-0002-2201-9402} \\
Department of Computer Science \& Engineering \\
SRM University AP, Amaravati, India \\
\href{mailto:ashokkumar.p@srmap.edu.in}{\texttt{ashokkumar.p@srmap.edu.in}}
\end{tabular}
&
\begin{tabular}{c}
\orcidlink{Swetha Ghanta}{0009-0005-5912-2138} \\
Department of Computer Science \& Engineering \\
SRM University AP, Amaravati, India \\
\href{mailto:swetha_ghanta@srmap.edu.in}{\texttt{swetha\_ghanta@srmap.edu.in}}
\end{tabular}
\end{tabular}
\end{center}

\vspace{0.15cm}

\begin{center}
\begin{tabular}{c}
\orcidlink{Laavanya Rachakonda}{0000-0002-7089-9029} \\
Department of Computer Science \\
University of North Carolina, Wilmington, USA \\
\href{mailto:rachakondal@uncw.edu}{\texttt{rachakondal@uncw.edu}}
\end{tabular}
\end{center}

\vspace{0.15cm}

\begin{center}
\begin{tabular}{c}
\orcidlink{Saraju P Mohanty}{0000-0003-2959-6541} \\
Department of Computer Science \& Engineering \\
University of North Texas, Denton, USA \\
\href{mailto:saraju.mohanty@unt.edu}{\texttt{saraju.mohanty@unt.edu}}
\end{tabular}
\end{center}

\begin{center}
\text{September 29, 2024} \\

\end{center}

\begin{abstract}
Maintaining excellent health and fitness depends on Maintaining health and fitness through a balanced diet is essential for preventing non-communicable diseases such as heart disease, diabetes, and cancer. NutriVision combines smart healthcare with computer vision and machine learning to address the challenges of nutrition and dietary management. This paper introduces a novel system that can identify food items, estimate quantities, and provide comprehensive nutritional information. NutriVision employs the Faster Region-based Convolutional Neural Network (Faster R-CNN), a deep learning algorithm that improves object detection by generating region proposals and then classifying those regions, making it highly effective for accurate and fast food identification even in complex and disorganized meal settings. Through smartphone-based image capture, NutriVision delivers instant nutritional data, including macronutrient breakdown, calorie count, and micronutrient details. One of the standout features of NutriVision is its personalized nutritional analysis and diet recommendations, which are tailored to each user’s dietary preferences, nutritional needs, and health history. By providing customized advice, NutriVision helps users achieve specific health and fitness goals, such as managing dietary restrictions or controlling weight. In addition to offering precise food detection and nutritional assessment, NutriVision supports smarter dietary decisions by integrating user data with recommendations that promote a balanced, healthful diet. This system presents a practical and advanced solution for nutrition management and has the potential to significantly influence how people approach their dietary choices, promoting healthier eating habits and overall well-being. This paper discusses the design, performance evaluation, and prospective applications of the NutriVision system.
\end{abstract}


\section{Introduction}
Indeed, the saying "wealth is nothing without health" underscores the pivotal role of nutrition in our overall well-being. Nutrients derived from food are essential for the proper functioning of our bodies, and deviations from a balanced diet can lead to various health issues. Fig. 1 illustrates the critical impact of both overeating and undernutrition on the development of chronic illnesses such as diabetes, heart disease, kidney disease, hypertension, obesity, and even life-threatening conditions like cancer (\cite{b52},\cite{b53}). Obesity and inflammation emerge as primary catalysts for many diseases, and the consumption of certain food elements, including color additives, chemicals, trans fats, refined sugar, salt, and processed foods, is linked to these health concerns (\cite{b54}). Consequently, individuals are modifying their dietary habits to address these challenges, prompting a widespread focus on diet regulation. Achieving a balance between dietary intake and monitoring is crucial for effective diet management.
\begin{figure}[h]
    \centering
    \includegraphics[width=0.7\linewidth]{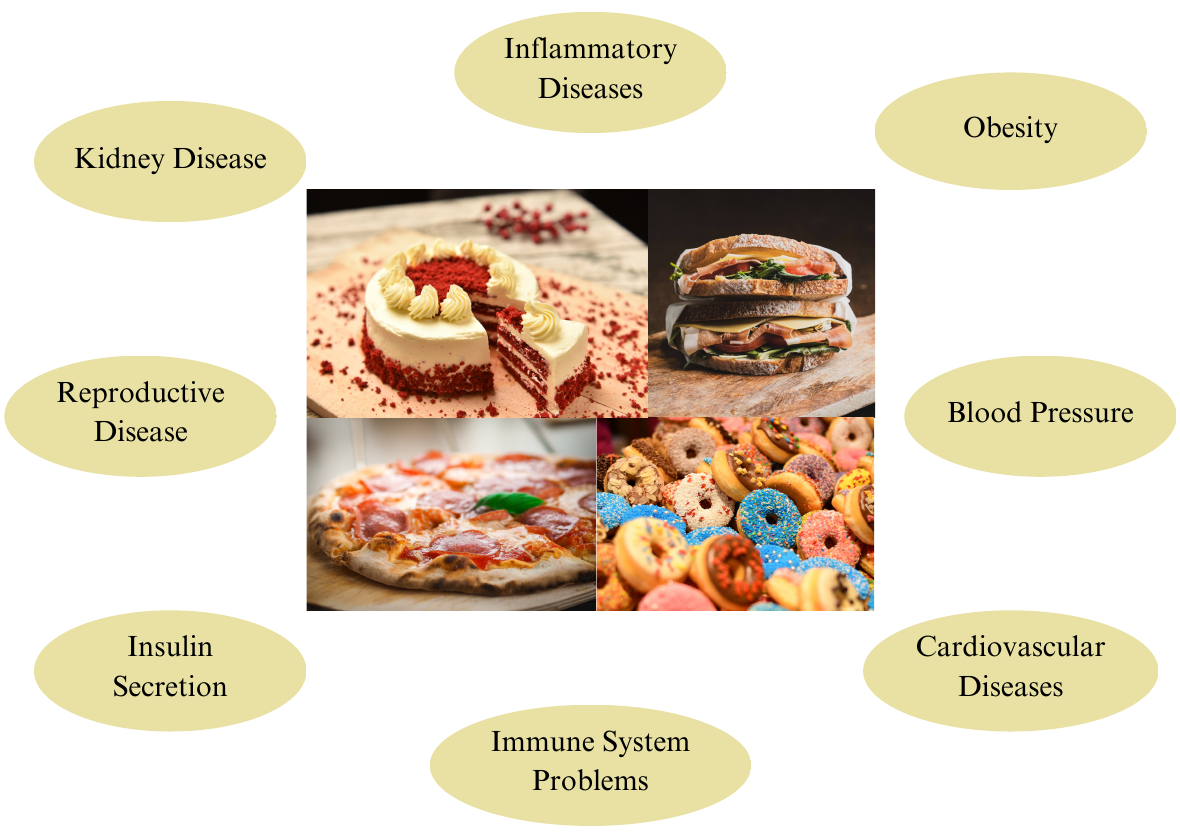}
    \caption{Impact of Unbalanced Diet}
    \label{fig:enter-label}
\end{figure}
\\Despite the proliferation of nutritional tracking devices in the market, many rely heavily on user input, necessitating manual entry of food intake (\cite{b55},\cite{b56}). This manual entry process can be cumbersome and time-consuming, leading to users hesitating to utilize these systems consistently. Additionally, dependence on user input raises the risk of inaccurate data entry, resulting in unreliable results. Commonly, these systems provide data on calorie quantities in food, consumption levels, and remaining allowances for the day, aligning with their primary goals of promoting a healthy weight and fitting into a particular diet framework. However, a crucial point emerges: being healthy doesn't always equate to adhering strictly to a low-calorie diet. A holistic approach to health considers various factors beyond mere calorie counts, emphasizing the importance of balanced nutrition and overall lifestyle choices.
\\\\In response to these challenges, there is a growing need for innovative solutions that offer accurate nutritional tracking without placing undue burdens on users, fostering sustained adherence and promoting a more comprehensive understanding of healthful dietary practices (\cite{b57}). Individuals who have high blood pressure ought to stick to a low-sodium diet, diabetics need to avoid sugar, and so on. Thus, monitoring sugar, sodium, saturated fat, protein, carbs, and other nutrients is crucial.
\\\\The research presented here explores the intersection of personalized healthcare, nutritional science, and computer vision, presenting a comprehensive system known as "NutriVision". The goal of the NutriVision system is to address the growing demand for a more informed and health-conscious approach to food selection. Essentially, NutriVision recognizes and detects food ingredients in user-provided photos using image recognition and AI-driven algorithms. After that, it embarks on a complicated journey of nutritional evaluation, quantifying the macronutrients, vitamins, and minerals present in the foods it has identified. However, what truly sets NutriVision apart is not just its capacity to deliver an accurate nutritional analysis but also its capacity to deliver personalized dietary suggestions and health guidance according to each user's unique health reports and goals.
\\Moreover, the adaptability of the NutriVision system to accommodate various dietary preferences and restrictions ensures that it can serve a wide audience, including vegetarians, vegans, and individuals with food allergies. By providing customizable options, NutriVision empowers users to make choices that align with their personal health needs and lifestyle preferences. This versatility not only enhances user engagement but also promotes a more inclusive approach to nutrition, ultimately fostering a supportive environment for diverse dietary practices.
\begin{figure}[h]
    \centering
    \includegraphics[width=1\linewidth]{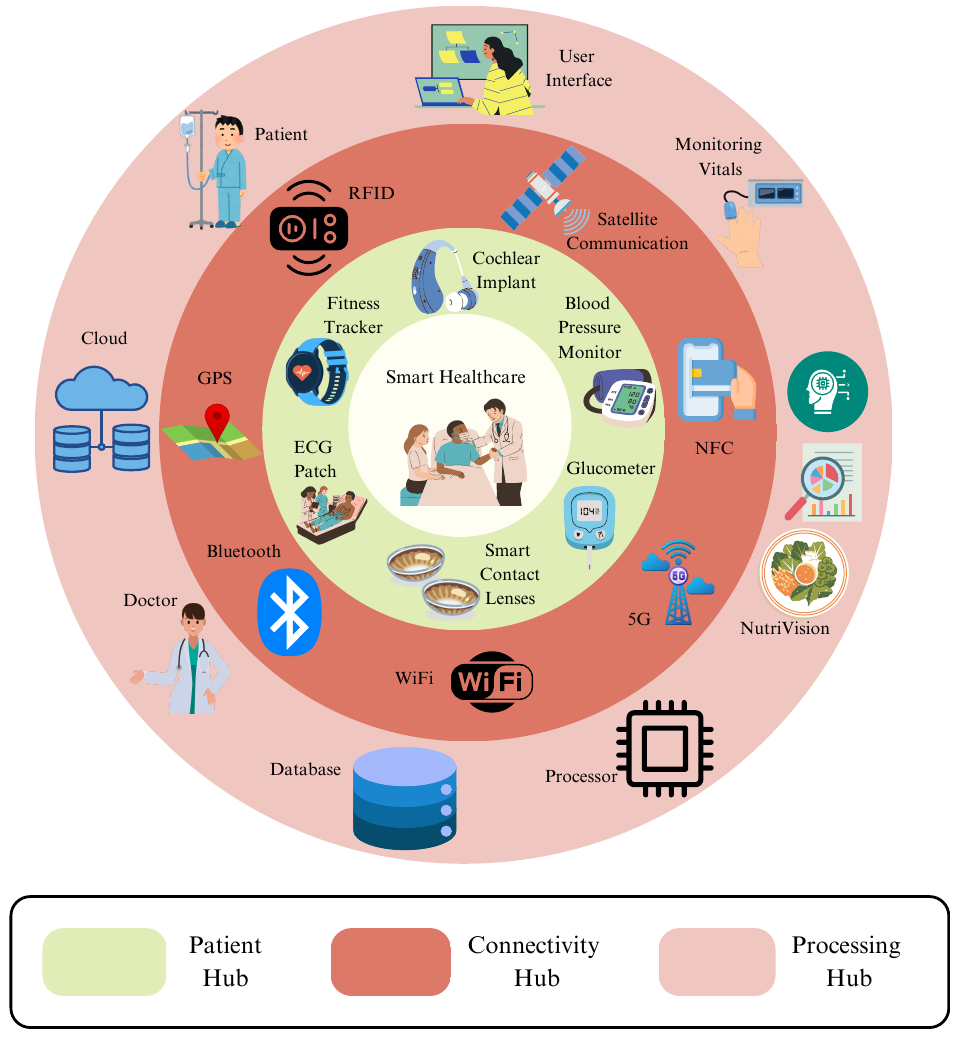}
    \caption{Smart Healthcare as a Healthcare Cyber-Physical System (H-CPS)}
    \label{fig:enter-label}
\end{figure}
\\Fig. 2 illustrates the Healthcare Cyber-Physical System (H-CPS) as a multi-layered model integrating various devices and technologies to monitor and enhance patient health in real time. At the core is the Patient Hub, comprising devices like fitness trackers, glucometers, and blood pressure monitors that directly track vital signs. Surrounding it, the Processing Hub includes AI algorithms, cloud computing, and data processors that analyze and interpret the collected data. NutriVision integrates within this hub as a nutritional management system, utilizing image recognition and machine learning to assess food intake and provide personalized dietary guidance.
\\\\NutriVision’s role in H-CPS focuses on nutrition as a key element of preventive healthcare. By offering real-time insights into food choices, it helps users make healthier decisions tailored to their unique health needs. The Connectivity Hub ensures smooth communication between devices via 5G, Wi-Fi, and other networks, enabling NutriVision to provide instant feedback and seamlessly integrate its recommendations into broader healthcare monitoring. Through this smart integration, NutriVision supports healthier eating habits, reducing the risk of diet-related diseases and contributing to overall health and wellness within the smart healthcare system.
\\\\As society increasingly recognizes the importance of nutrition in disease prevention and overall health, tools like NutriVision stand to play a transformative role in empowering individuals to make informed dietary choices. By leveraging advanced technologies such as image recognition and machine learning, NutriVision not only simplifies the process of tracking nutritional intake but also enhances users' awareness of their eating habits. This intuitive approach fosters a deeper connection between individuals and their food choices, encouraging them to embrace healthier eating patterns. Furthermore, by providing personalized insights and recommendations tailored to specific health conditions, NutriVision aims to bridge the gap between dietary knowledge and practical application, ultimately supporting users in achieving their health goals and improving their quality of life. This integration of technology and nutrition is poised to contribute significantly to public health initiatives aimed at reducing diet-related diseases.
\\\\This is how the other parts of the paper are arranged.  The innovative aspects of the suggested solution are discussed in Section \ref{sec:novel_contributions}. We review several related works and provide background information for our work in Section \ref{sec:related_work}. Together with the test findings, a thorough explanation of the recommended solution is given in Section \ref{sec:framework}. Section \ref{sec:validation} contains the results and some comparative analysis. The paper is concluded in Section \ref{sec:conclusion} with a discussion of upcoming projects.

\section{Novel Contributions of the Current Paper}
\label{sec:novel_contributions}
\subsection{Problem Addressed}
Current food nutrition monitoring systems often lack personalization and real-time estimation capabilities. Users frequently face challenges with generic dietary advice that doesn’t account for individual health data. Additionally, there is a need for systems that provide immediate feedback on the nutritional content of food and tailor recommendations based on daily nutrient requirements.
\subsection{Solution Proposed by NutriVision}
NutriVision addresses these issues with a state-of-the-art system that offers several significant advantages. First, it provides automated, real-time nutritional content estimation, allowing users to understand the nutritional value of their food before consumption. This feature ensures that users receive instant feedback on their meals, enhancing their awareness of nutritional intake. Second, NutriVision integrates user-specific health data to deliver personalized dietary recommendations tailored to individual needs, based on metrics such as BMI and dietary preferences. The system also considers daily nutrient intake requirements when suggesting meals, ensuring that advice is relevant and promotes a balanced diet. By eliminating the need for manual entry of food data, NutriVision streamlines the monitoring process, reducing user effort and minimizing errors. Additionally, the system's real-time adaptation feature allows it to adjust recommendations based on continuous feedback and data, providing ongoing improvement in dietary guidance. This combination of real-time data, personalization, and adaptability enhances user convenience, making NutriVision an ideal solution for managing nutritional intake with minimal effort and maximum efficiency.
\subsection{Significance of the Proposed Solution}
NutriVision’s innovative approach revolutionizes food nutrition monitoring by offering an interactive platform that improves dietary habits. Its real-time estimation and personalized recommendations provide users with actionable insights that are directly applicable to their daily lives. The system’s ability to adapt recommendations based on feedback ensures continuous improvement and relevance. This combination of real-time data, personalization, and adaptability makes NutriVision an ideal solution for those seeking to enhance their dietary health with minimal effort and maximum efficiency.

\section{Related Work}
\label{sec:related_work}
In light of the contemporary surge in advanced hardware technologies, particularly in high-performance computing processors and devices, coupled with the evolution of diverse computing paradigms such as the Internet of Things (IoT), edge-based computing platforms, and tailored artificial intelligence (AI) models for edge computing, innovative models like smart healthcare and intelligent agriculture have materialized into practical applications. Notably, the realm of automatic dietary intake estimations has witnessed notable advancements, predominantly manifesting through mobile app-centric solutions. This segment delves into pertinent research endeavors within this domain.
\\\\A systematic classification of diet monitoring approaches is shown in Fig. 3, which differentiates between automated and manual techniques. Manual diet monitoring methods require individuals to actively engage in tracking their food intake, with users being responsible for recording their dietary habits. One of the simplest forms is maintaining a food journal, where individuals document what they eat throughout the day, either on paper or digitally through spreadsheets or text files. This approach heavily relies on user memory and honesty, making it prone to inconsistencies. In recent years, various smartphone apps have emerged to streamline this process, offering a digital interface for meal input and integrating nutritional databases to estimate caloric intake and macronutrient breakdowns.
\\\\Calorie counting is another widely used manual approach, involving the calculation of the caloric value of consumed foods through nutritional labels, online resources, or databases. While some apps assist with this, users remain responsible for ensuring the accuracy of their entries, introducing potential for human error. Issues such as overestimating portions or misidentifying food types can lead to inaccurate caloric calculations, which may derail dietary goals. These manual methods are often time-consuming and heavily dependent on user motivation, discipline, and consistency. This reliance on active participation increases the likelihood of under- or over-reporting, as human error—such as forgetting to log certain meals or incorrectly estimating portion sizes—significantly affect the results of manual diet monitoring.
\begin{figure}[h]
    \centering
    \includegraphics[width=1\linewidth]{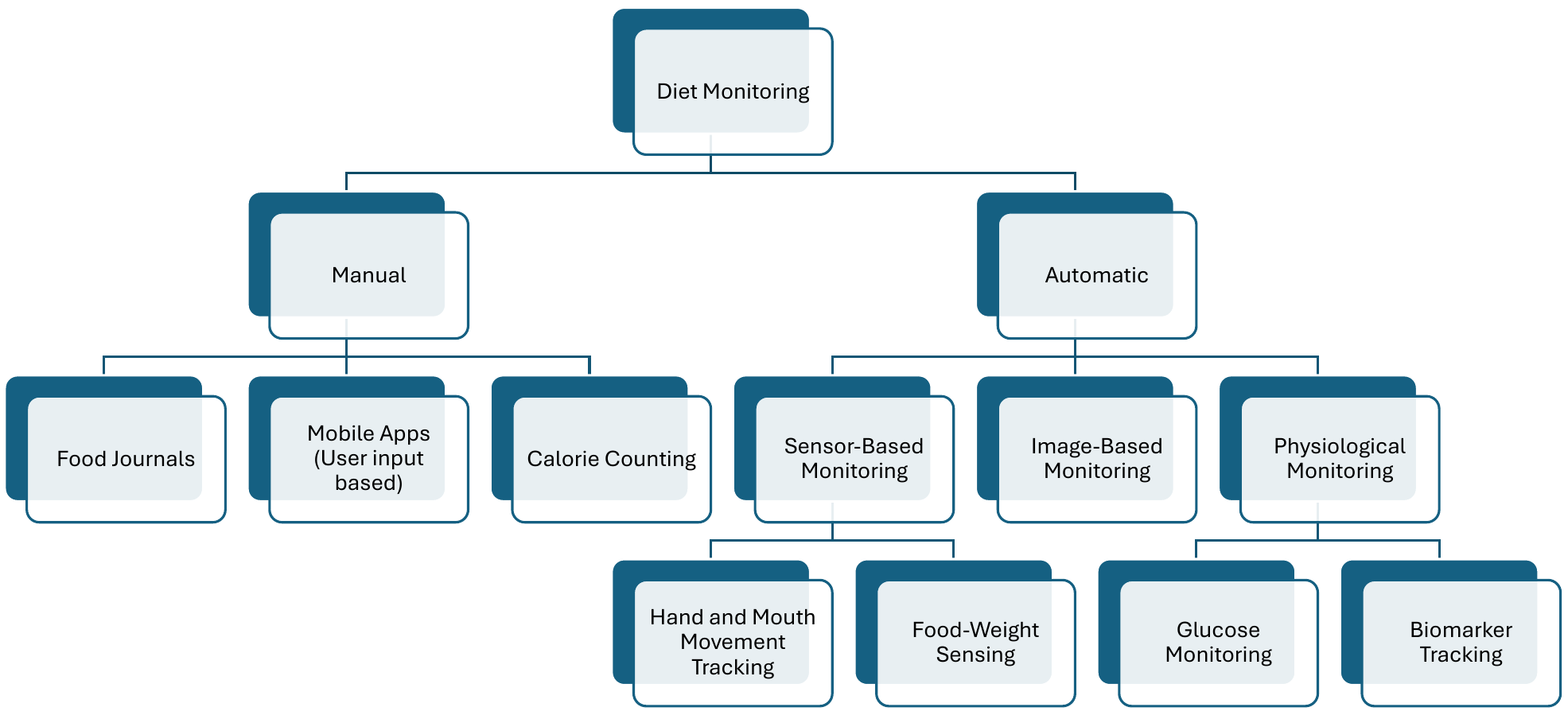}
    \caption{Diet Monitoring Approaches}
    \label{fig:enter-label}
\end{figure}
\\\\On the other hand, automated approaches utilize cutting-edge technology to reduce user burden, making the process more efficient, precise, and less prone to human error. One key approach in the automated category involves sensor-based systems. For instance, food-weight sensing utilizes smart plates, bowls, or scales that can automatically quantify the weight of the food. These systems are often integrated with AI to recognize food types and calculate nutritional values. Another type of sensor-based approach tracks hand-to-mouth movement through wearable devices like smartwatches or smart glasses. These wearables detect movement patterns and infer eating episodes by monitoring hand gestures and bite counts. This method can automatically log meals and estimate intake frequency, reducing reliance on self-reporting.
\\\\\\Another advanced technique for automated diet tracking is image-based monitoring. This approach leverages artificial intelligence (AI) and image recognition technologies to analyze photos of food and estimate the portion size and nutritional content. Users simply take a picture of their meal, and AI algorithms trained on large food datasets identify the food items and generate estimates for calories, macronutrients, and portion sizes.
\\\\In addition to visual and sensor-based methods, physiological monitoring offers another layer of precision in automated diet tracking. These techniques include continuous glucose monitoring (CGM) devices and biomarker analysis. For example, CGM systems monitor blood sugar levels in real-time, providing insights into how specific foods impact an individual’s glucose levels. Similarly, biomarker tracking analyzes bodily fluids such as saliva, blood, or urine to detect key nutritional markers. These physiological indicators help to track metabolic responses and offer a more comprehensive picture of how the body processes consumed foods.
\\\\An edge-cloud procedure has been used and food in the cloud is classified using a convolutional neural network (CNN) in \cite{b1}. Although food has been identified, this work has not estimated nutritional value. The user is informed whether or not he is stress eating, and some exercise is advised. Only the number of calories has been estimated. It does not give any personalized diet advice to the users. While in \cite{b2}, 19 classes of Food-101 dataset has been identified. The user does not need to provide any kind of manual input , given it is fully automated.  It uses a CNN(convolutional neural network) approach for food detection. It provides estimation of nutrients and not just calories. It has difficulty in handling customized food orders. But any kind of personalized advice and suggestions are not given.
\\\\An edge-cloud method for dietary assessment has been used in \cite{b8}. The food images have been processed at the edge devices using image textural features, and the food in the cloud is classified using a convolutional neural network (CNN). Despite the fact that food has been identified in the Food-101 dataset, no nutritional value has been estimated in this work. Real-time estimations of food attributes are presented in \cite{b9}. A textual corpus has been employed for the identification of food attributes, integrating deep learning methodologies for the recognition of food items. While the context of our work is the same, the nutrients are estimated based on the image. 
\\\\Food nutrients were analyzed and calories were estimated using mask-RCNN and union post processing in \cite{b10}. The estimation of food weight involves the utilization of both the pixel count within the mask and the implementation of a linear regression model. Food serving sizes and their nutritional and calorific contents are estimated in \cite{b20}. There has been no quantification in this case.
\\\\A deep autoencoder network has been used to determine the vitamin A content based on the color of the pureed food in \cite{b21}. Only one particular nutrient has been found in this work. To detect food for diabetic patients, researchers have implemented the bag-of-features model, as referenced in \cite{b18}. It calculates dense local features in the HSV color space through scale-invariant feature transforms to generate a visual dictionary consisting of 10,000 words. Subsequently, food photos are classified using linear support vector machine classifiers. Here, no nutrient value is computed. In a different investigation, the Food-Ingredient Joint Learning module was employed for ingredient recognition, and the Attention Fusion Network was utilized for fine-grained food recognition, as documented in \cite{b22}. Although the accuracy in food recognition is substantial, the accuracy in ingredient recognition varies from moderate to low.
\\\\For ingredient recognition and food classification, a deep CNN has been employed \cite{b19}. However, this work does not show any nutritional value. \cite{b5} is dispersed across multiple computer platforms. There is a suggested food calorie estimation app that is mobile and cloud-based. Here, the food is identified using a cloud-based Support Vector Machine using the MapReduce technique. Calorie counting is done both before meals and after eating leftovers. For that, image graph cut segmentation has been employed. To establish a reference point for measurement, it is essential to have a thumb present within the frame of the food photo. The calorific value has also been determined here. Variations in thumb size are another problem. The person's height affects the thumb size, which can lead to inaccurate results.
\\\\The work \cite{b11} introduces a model for food recognition based on Convolutional Neural Networks (CNN) and utilizes text2vec for attribute estimation. The model is client-server based. But the system does not include cooked foods, mixed foods, or liquid foods. The calorific value of the food is the main focus here as well. A wearable system based on piezoelectricity was used in \cite{b23} to identify food and calculate caloric intake. When eating, skin movement from the lower trachea is detected by the piezoelectric sensors. The food's weight was determined, and then the final calorie count was carried out. Here, no additional nutritional value has been tallied.
\\\\Diverse methodologies have been documented across several publications for determining calorific values from food volume. Notably, a multilayer perceptron model is presented in \cite{b24}, an image analysis-based approach is outlined in \cite{b25}, and a CNN-based method is detailed in \cite{b26}. Additionally, some works, including \cite{b27}, extend their focus beyond calorific values to encompass ingredient lists and cooking instructions, employing multitask CNN for this purpose. In a distinct context, the work \cite{b28} categorizes various food types. This observation underscores that, while numerous papers address the computation of calorific values, a comparatively limited number delve into the broader spectrum of food nutritional value.
\\\\Several studies have utilized various methodologies for dietary intake estimation. In the work \cite{b45}, an innovative solution using inertial sensors was introduced. This system tracks eating gestures and combines them with image-based food identification algorithms to estimate calorie intake. However, it lacks real-time feedback and nutrition advice tailored to the user. Similarly, \cite{b42} employed a multimodal approach that integrates visual and acoustic data for food intake monitoring, though this system also focuses solely on calorific estimation rather than complete nutritional analysis. 
\\\\In an effort to automate dietary tracking, \cite{b49} proposed a smartphone-based system that detects food using deep learning models. This system emphasizes ease of use, eliminating the need for user input, but suffers from accuracy issues when it comes to complex food mixtures. Nutritional value estimation is limited to caloric content, with no focus on personalized diet suggestions. 
\\\\In \cite{b48}, a cloud-based mobile app that allows users to capture images of their meals is used. The food recognition is powered by CNNs, and the app estimates macronutrient content. However, this approach is limited by its reliance on cloud resources, which might hinder its performance in low-connectivity environments. Another study \cite{b46} explored an AI-driven mobile app that utilizes food diaries and manual data entry to track food intake. While this tool offers a holistic nutritional breakdown, it lacks automation, as users must input data manually. 
\\\\Other researchers have focused on ingredient recognition as an intermediary step toward dietary analysis. For instance, \cite{b47} introduced a model that uses a CNN to recognize ingredients in complex dishes. Their work emphasizes ingredient classification rather than a complete nutritional breakdown. Similarly, \cite{b39} employed a CNN-based approach for ingredient recognition in mixed dishes, but like \cite{b47}, their system struggles with accurate nutritional estimation. 
\\\\A unique contribution is made by \cite{b44}, where a system using a graph-based approach to food recognition is discussed. This method segments the food image into portions and then uses those segments for caloric and nutritional estimation. However, the system is prone to inaccuracies when dealing with complex meals. \cite{b50} tackled this issue by proposing a method that integrates portion size estimation with ingredient recognition, though their model is still primarily concerned with calorie counting rather than a full nutritional assessment. 
\\\\In \cite{b38}, a hybrid system that combines image analysis with manual data entry for diet tracking is introduced. Although effective in estimating calories and basic nutrients, it lacks the level of automation seen in other state-of-the-art systems. Another notable mention is the research conducted by \cite{b43}, which explored deep learning algorithms for estimating the nutritional value of food in a fully automated way. However, the model was trained on a limited dataset, and its real-world application may be hindered by a lack of diversity in the food items it can recognize. 
\\\\In a similar way, \cite{b41} presented a dietary tracking system that uses a knowledge-based approach, drawing on food ontologies to estimate nutrients and suggest personalized meal plans. While innovative, the system relies heavily on an external database, which may not cover all local food variations. Another edge-cloud system designed by \cite{b51} focused on identifying food in real-time but offered limited nutritional data, focusing primarily on calorie estimation. The study conducted by \cite{b40} uses a mask-RCNN approach to segment food images and estimate portion sizes for caloric calculation. While their system is effective for identifying meal portions, it does not extend to broader nutritional analyses, leaving a gap in personalized dietary recommendations.
\\\\Table 1 offers an overview of several studies for a more comprehensive context, highlighting significant results and insights from a variety of research work. By comparing these systems side by side, the table underscores the unique contributions of NutriVision, particularly in areas where other systems fall short, such as nutritional analysis and the provision of an interactive platform.  
\begin{table}[h]
    \centering
    \caption{Comparative Analysis of Self-Tracking Systems}
    \label{tab:comparative_analysis}
    \renewcommand{\arraystretch}{1.5}
    \begin{tabularx}{\textwidth}{|X|p{1cm}|X|p{1.8cm}|p{1.5cm}|p{1.8cm}|p{1.5cm}|}
        \hline 
        \textbf{Tracking System} & \textbf{Input} & \textbf{Analysis} & \textbf{Fully-Automated System} & \textbf{Nutritional Value Estimation} & \textbf{Personalized Advice} & \textbf{Interactive Platform} \\
        \hline\hline
        Harrison, et al. (2010) \cite{b29} & Image & Not Feasible & No & No & No & No \\
        \hline
        O. Beijbom, et al. (2015) \cite{b30} & Image & Not Entirely Possible to be Done & Semi-Automated & No & No & No \\ 
        \hline
        Jiang, et al. (2018) \cite{b31} & Image & Not Very Beneficial & Semi-Automated & No & No & No \\
        \hline
        Pouladzadeh, et al. (2015) \cite{b34} & Image & Not Very Beneficial & Semi-Automated & No & No & No \\
        \hline
        L. Rachakonda, et al. (2019) \cite{b35} & Image & Yes & Semi-Automated & No & No & No \\
        \hline
        Taichi, et al. (2009) \cite{b32} & Image & No Nutritional Info & No & No & No & No \\
        \hline
        L. Rachakond, et.al (2020) \cite{b1} & Image & Yes, Calorie Estimation and Stress Detection & Yes & No & No & No \\
        \hline
        A. Mitra, et al. (2022) \cite{b2} & Image & Yes, Nutrition Estimation & Yes & Yes & No & No \\
        \hline
        M.-L.Chian, et al. (2019) \cite{b10} & Image & Yes, Nutritional Estimation & Semi-Automated & Yes & No & No \\
        \hline
        P. Pouladzadeh, et al. (2014) \cite{b5} & Image & Yes, Calorie Estimation & Semi-Automated & No & No & No \\
        \hline
        \textbf{NutriVision (Current-Paper)} & \textbf{Image} & \textbf{Yes, Nutrition Estimation and Personalized Nutrition Analysis} & \textbf{Yes} & \textbf{Yes} & \textbf{Yes, Food recommendations based on health history and BMI.} & \textbf{Yes, it answers diet and health queries.} \\
        \hline
    \end{tabularx}
\end{table}

\section{The Innovative Framework: NutriVision}\label{sec4}
\label{sec:framework}
\subsection{Framework}

This section details the intricacies of food nutritional value estimation and personalization, mirroring the process illustrated in Fig. 4. The user leverages her smartphone to capture an image of the food alongside a reference object, facilitating the determination of the nutritional value of the food. In this section, comprehensive details regarding the reference object are expounded. Subsequent to capturing the image, features specific to the food are extracted and employed for food detection. The nutritional value is then determined by comparing the identified food object with entries in a nutrition database, representing average serving sizes. Discrepancies between the serving size and stored values may exist, necessitating food quantification. Hence, the reference object plays a crucial role. Following the quantification process outlined in this section, the nutritional values are presented. Fig. 5 illustrates the developmental workflow of NutriVision, spanning from dataset collection to inference, showcasing the comprehensive journey of the system. This workflow highlights the importance of accuracy in both image processing and database comparisons, as these steps are vital for ensuring reliable nutritional assessments. \\
\begin{figure}[h]
    \centering
    \includegraphics[width=0.32\linewidth]{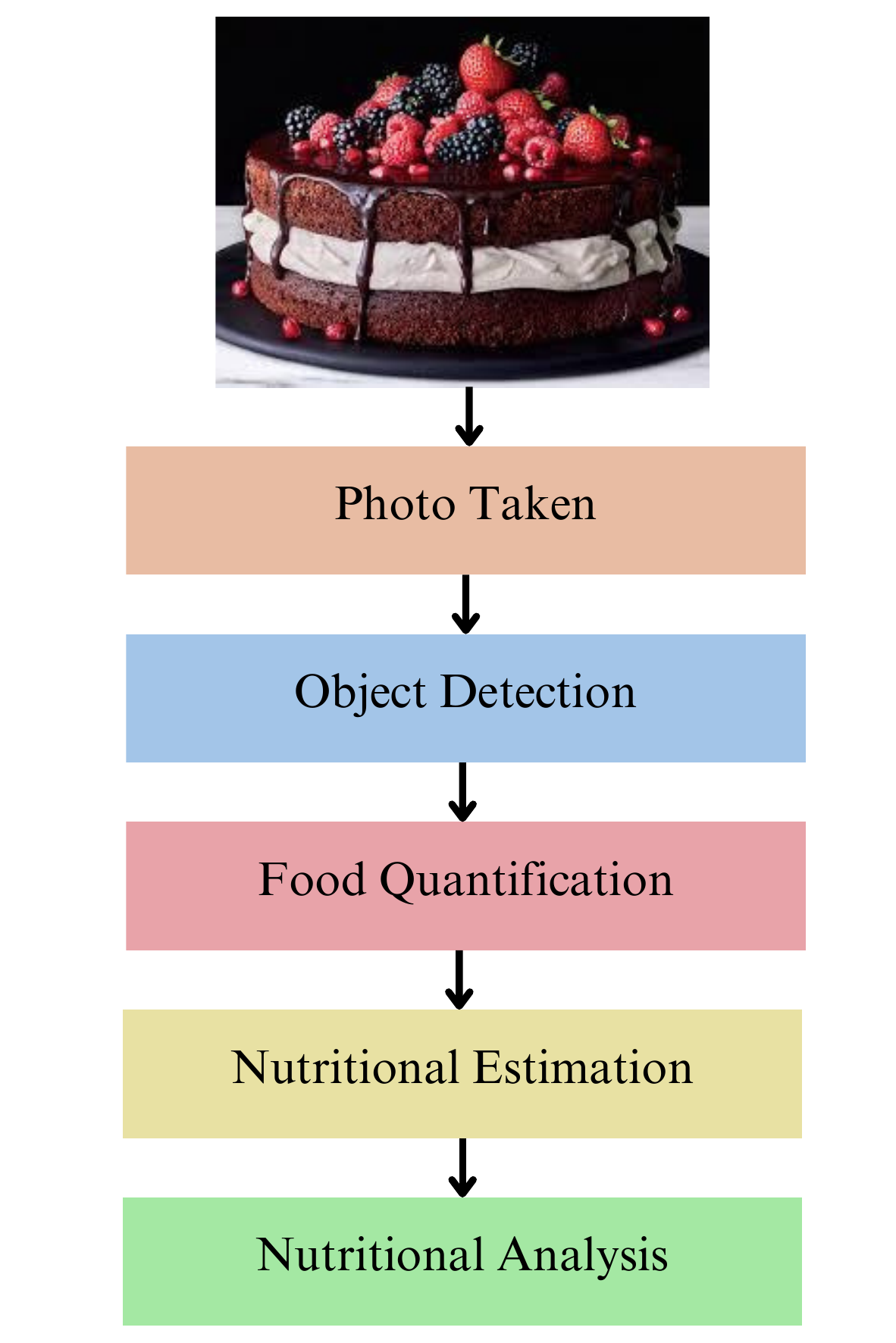}
    \caption{Process Workflow of NutriVision}
    \label{fig:enter-label}
\end{figure}
\begin{figure}[h]
    \centering
    \includegraphics[width=0.85\linewidth]{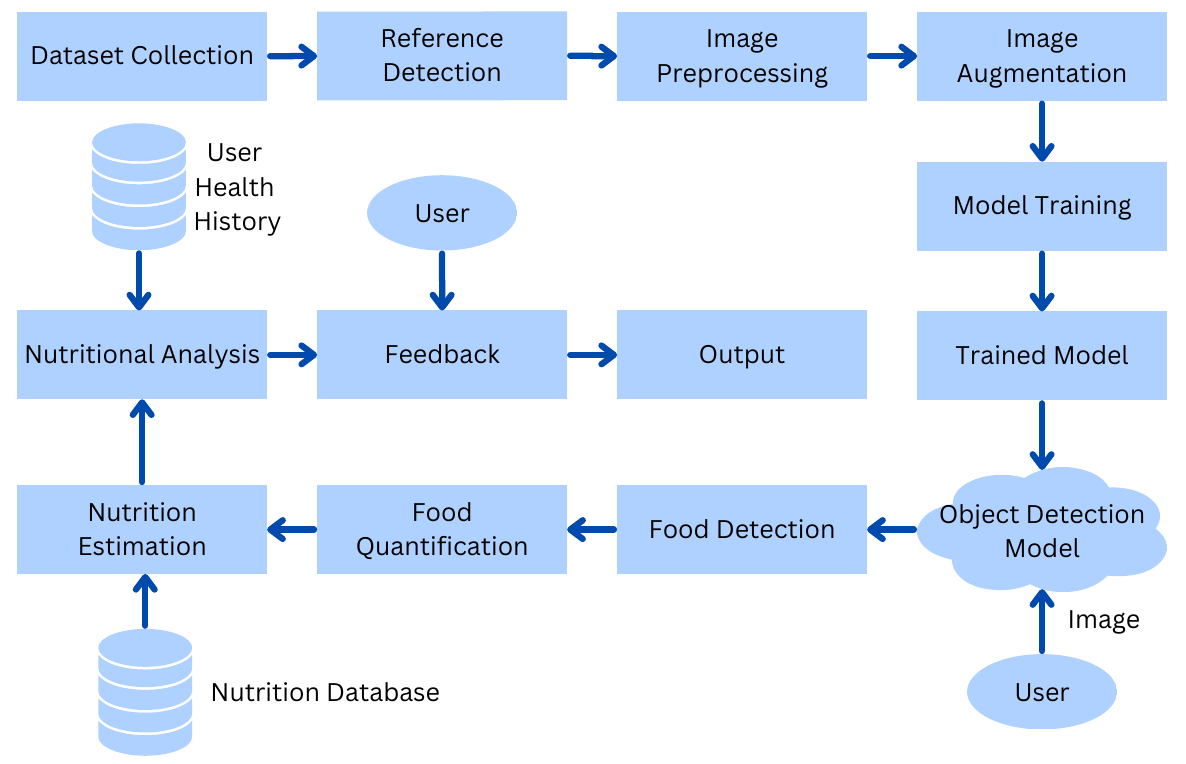}
    \caption{Development Workflow of NutriVision}
    \label{fig:enter-label}
\end{figure}
\\
\subsubsection{Acquisition of Dataset} Food classification employs a tailored and customized dataset, incorporating images sourced from the COCO Dataset\cite{b15} and customized dataset \cite{b14}. In contrast to COCO's extensive set of 91 classes, our customized dataset is streamlined to encompass a total of 10 specific classes. The annotation of images with bounding boxes is conducted through Roboflow \cite{b33}. The dataset comprises a total of 500 images, with 450 designated for training and 50 for testing, encompassing the 10 defined classes: Apples, Oranges, Pizza, Broccoli, Cake, Donut, Sandwich, Carrot, Banana, and Hotdog. It is noteworthy that, for addressing nutritional aspects, a distinct custom dataset has been employed.
\subsubsection{Reference Detection} In food quantification, a critical consideration is the selection of a reference for accurate measurement. The size of the food plate in a given picture can vary based on the proximity of the smartphone camera to the table during photo capture. Quantifying food can be achieved through various methods, including measuring the distance between the plate and the phone, employing image segmentation, or utilizing a reference object. For simplicity, we opt for the third method, involving the use of a reference object, in this instance. Specifically, a one rupee coin with a diameter of 21.93 millimeters is employed as the reference given in Fig. 6. Positioned on the same table next to the plate, the coin and the food are captured within the same frame during photo capture. The detection of the coin, facilitated by Algorithm 1, establishes it as a reference. As outlined in Section \ref{sec:validation}, the portion size of the identified food is determined by leveraging the dimensions of the detected coin.
\begin{algorithm}[h]
\caption{Detection of Reference Coin}
\label{algo1}
\SetAlgoNlRelativeSize{0} 
\SetAlgoNlRelativeSize{-1}
\SetAlgoNlRelativeSize{-2}
\begin{enumerate}
    \setlength{\itemindent}{-1em} 
    \item Colour of the coin is set.
    \item The image is transformed from RGB to HSV colour space.
    \item The area of the coin is calculated.
    \item Determine the ratio between the image and the coin's actual sizes.
    \item Locate all the contours.
    \item Coin's contour is selected.
    \item Minimum Bounding Box is determined.
    \item Note the Bounding Box's Length and Width.
\end{enumerate}
\end{algorithm}
\begin{figure}[h]
    \centering
    \includegraphics[width=0.5\linewidth]{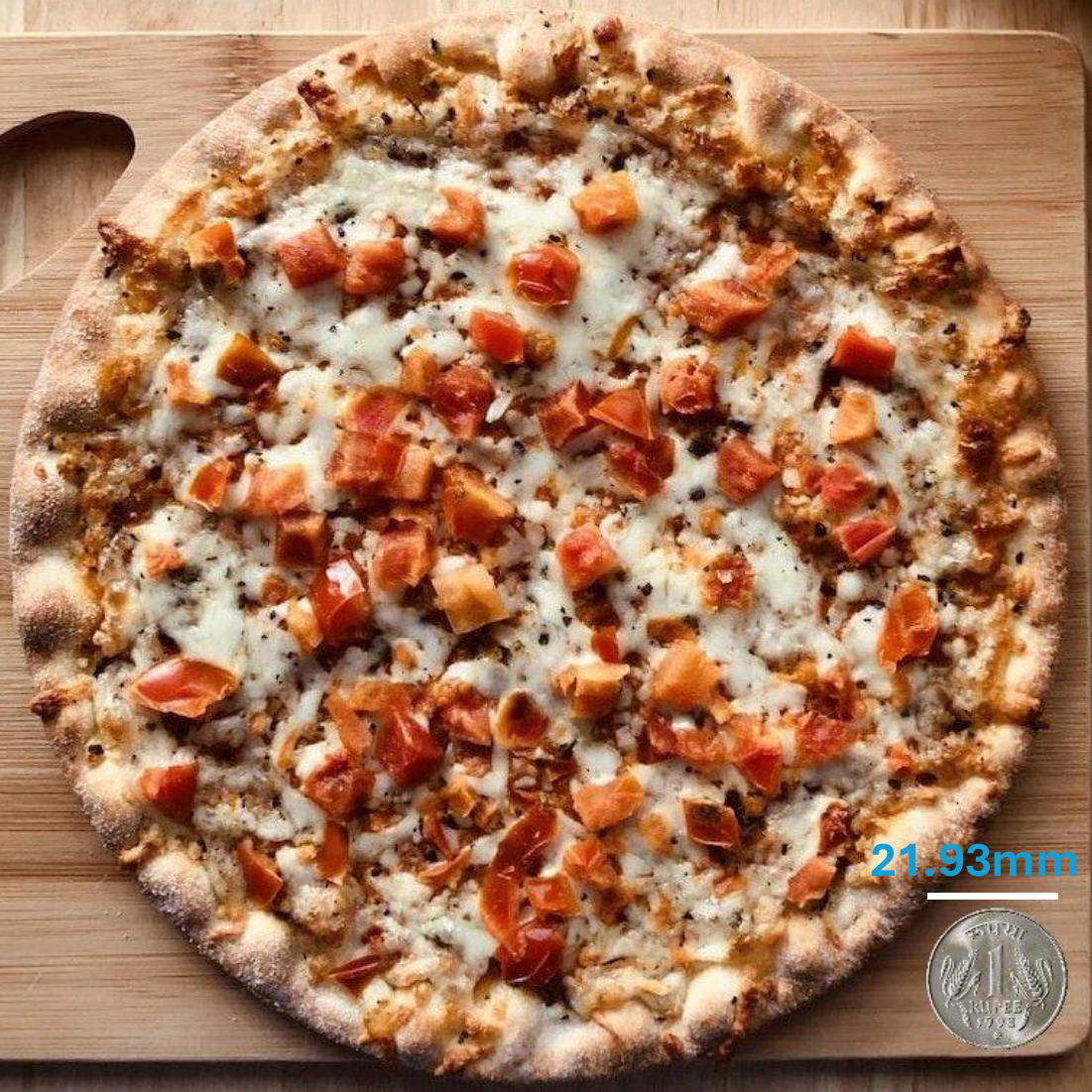}
    \caption{Reference Selection of Nutrivision}
    \label{fig:enter-label}
\end{figure}

\subsubsection{Image Preprocessing and Augmentation} The images have been subjected to preprocessing, involving size and normalization adjustments before undergoing data augmentation. In an effort to enrich the diversity of the training dataset, additional data was introduced. This involved cropping the images with a minimum scale size of 0.1 and a maximum scale size of 2.0, along with horizontal flipping. The setting for maximum dimension padding has been retained as True.

\subsubsection{Model Training} 
The feature extractor for the object detection model utilizes a pre-trained Faster R-CNN model \cite{b16} on the COCO dataset \cite{b7}. The application of transfer learning not only streamlined the training process but also significantly improved accuracy. Faster R-CNN is well-regarded for its effective object detection capabilities \cite{b17}, incorporating a Region Proposal Network (RPN) with a convolutional neural network (CNN) backbone, such as ResNet. The Region Proposal Network (RPN) within Faster R-CNN is integral to the model's functionality, handling the localization of objects. Specifically, the RPN is responsible for generating region proposals, representing potential bounding boxes around objects within the image. Subsequently, these region proposals contribute to the precise localization of objects. The collaborative operation of the RPN and the feature maps extracted from the ResNet-50 backbone \cite{b6} plays a crucial role in proposing regions of interest within the image.

\subsubsection{Food Quantification and Nutrition Estimation} After identifying the food and reference coin in the image, a comparison is made between the names of the food items in the nutritional database and the detected food label. The nutritional values from the dataset that have been stored correspond to predefined values for a specific portion size. Algorithm 2 is used to properly quantify the food that has been detected. \\\\
\begin{algorithm}[H]
\caption{Nutritional Value Estimation from Food Image}
\label{algo2}
\SetAlgoNlRelativeSize{0} 
\SetAlgoNlRelativeSize{-1}
\SetAlgoNlRelativeSize{-2}

\KwIn{Image to be tested}
\KwOut{Nutritional Value of the food}

\begin{enumerate}
    \item Take a picture of a plate of food from above with a one rupee coin next to it.
    \item To generate bounding boxes of recognized food on the plate, run the image through the food detection model.
    \item Use Algorithm 1 to measure the coin area in relation to the image.
    \item Determine the ratios between the 1 rupee coin's actual diameter (21.93 millimeters) and the identified coin's pixel diameter in the picture.
    \item To determine the true dimensions of the observed food, multiply the bounding box dimensions by the corresponding ratios.
    \item Use the food's preset generalized height based on the class of food that has been detected.
    \item Determine the identified food's total volume in cubic centimeters by calculating its length, width, and height.
    \item To take into consideration the additional area that comes from the rectangular boundary boxes, multiply this volume by a constant factor (such as 0.8).
    \item Calculate the number of grams of the particular food on the plate by using the cubic centimeter to gram converter.
    \item Lastly, use the nutrition database to derive the associated nutritional value.
\end{enumerate}
\end{algorithm}
Here, the food's actual dimension is determined by contrasting it with the object of reference. To account for the excess space in the rectangular bounding box, the computed volume is multiplied by 0.8. In the rectangular bounding box, there are empty spaces where there is no food because most of the food is served in circular or oval plates. Following the estimation of the number of pixels within the bounding box's empty space in comparison with the bounding box as a whole, the value of 0.8 was carefully selected and confirmed across multiple images. 
\\\\Finally, the nutritional values of the food on the plate are approximated by converting the volume into weight (using data from the Food-A-Pedia\cite{b12} database). After the calculation of the food quantity, the nutritional value is estimated by using the custom nutrition database which contains the nutrient information of different foods for a single serving (100 grams). The nutritional content of the detected food is calculated accordingly.

\subsubsection{Nutritional Analysis} In this research study, we use the latest technologies and sophisticated algorithms to present a technically sophisticated method to individualized nutritional analysis by integrating collaborative filtering approaches with content-based filtering techniques. Important information is extracted from consumers' health reports via the content-based filtering component using machine learning methods, notably natural language processing (NLP). 
\\\\\\A thorough user profile is created using this data, which also includes nutritional objectives, dietary limitations, and medical concerns. To convert text data into numerical vectors for the NLP model, we use the TF-IDF (Term Frequency-Inverse Document Frequency) vectorizer from the scikit-learn module. By allocating weights based on word frequency in relation to the complete document collection, TF-IDF efficiently captures the meaning of each word within the context of a given document. Each word in the user's health history correlates to a feature, and the history is handled like a document. After processing this text data, the TF-IDF vectorizer generates numerical vectors for each user that indicate the significance of different health-related phrases. Then, cosine similarity between these user vectors and the meal descriptions is computed in order to find the closest matches and produce customized food recommendations. We use matrix factorization and Singular Value Decomposition (SVD) methods for the collaborative filtering component to uncover latent patterns in user-item interactions. 
\\\\By using these methods, the algorithm may find minute relationships between food items and users, which helps to improve the recommendations even further. To guarantee effective model training and suggestion production, we include these algorithms into sophisticated programming frameworks like PyTorch or TensorFlow. This hybrid strategy combines the best features of both filtering techniques, enhancing the accuracy and applicability of dietary advice by combining cutting-edge technology and complex algorithms. 
\\\\An intelligent chatbot that converses with users to obtain information about their dietary preferences, past medical conditions, and food selections facilitates the entire process. Along with gathering this data, the chatbot also provides real-time, customized recommendations, according to user reaction and continued communication. 
\begin{figure}[h]
    \centering
    \includegraphics[width=0.6\linewidth]{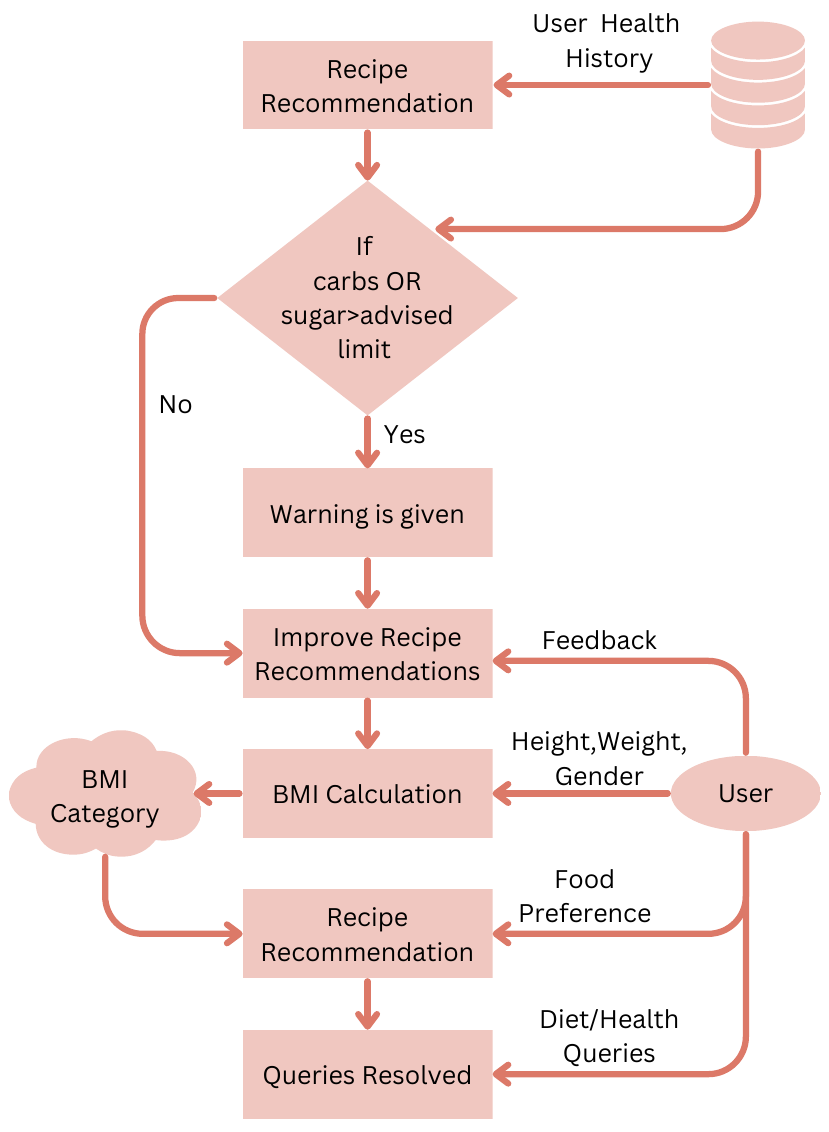}
    \caption{Workflow of Nutritional Analysis}
    \label{fig:enter-label}
\end{figure}
\\The nutritional analysis module's workflow is depicted in Fig. 7, which also emphasizes the chatbot's contribution to personalized recipe recommendations based on user health data. First, based on the user's medical history, a recipe recommendation is generated. A warning is sent if the suggested amounts of sugar or carbohydrates are higher than the safe thresholds. Following that, users offer feedback on the recommendations, allowing for ongoing improvement of the list. In order to update the suggestions in light of changing food patterns, the system also computes BMI. Some more recipe recommendations are given according to the BMI category and nutritional content of previously consumed meals. Lastly, users can ask the chatbot more questions to make sure their decisions support their health objectives. This technical connection demonstrates our commitment to offering a robust and cutting-edge personalized nutritional analysis solution, ensuring that users receive timely and accurate dietary recommendations tailored to their individual health needs.

\section{Validation of NutriVision}
\label{sec:validation}

\subsection{Validation}
A carefully chosen set of 200 photos was used in the model evaluation process to assess how well the categorization system worked; the outcomes are shown in Fig. 8. Even after localization, the model's ability to categorize food products has shown to be very accurate, producing exact results almost all of the time. The model classifies and recognizes various food products with remarkable accuracy. But the assessment also identified several difficulties. In particular, when the photos show food items next to one another with little space between them on the plate, it may compromise the accuracy of the model. 
\begin{figure}[h]
    \centering
    \includegraphics[width=0.67\linewidth]{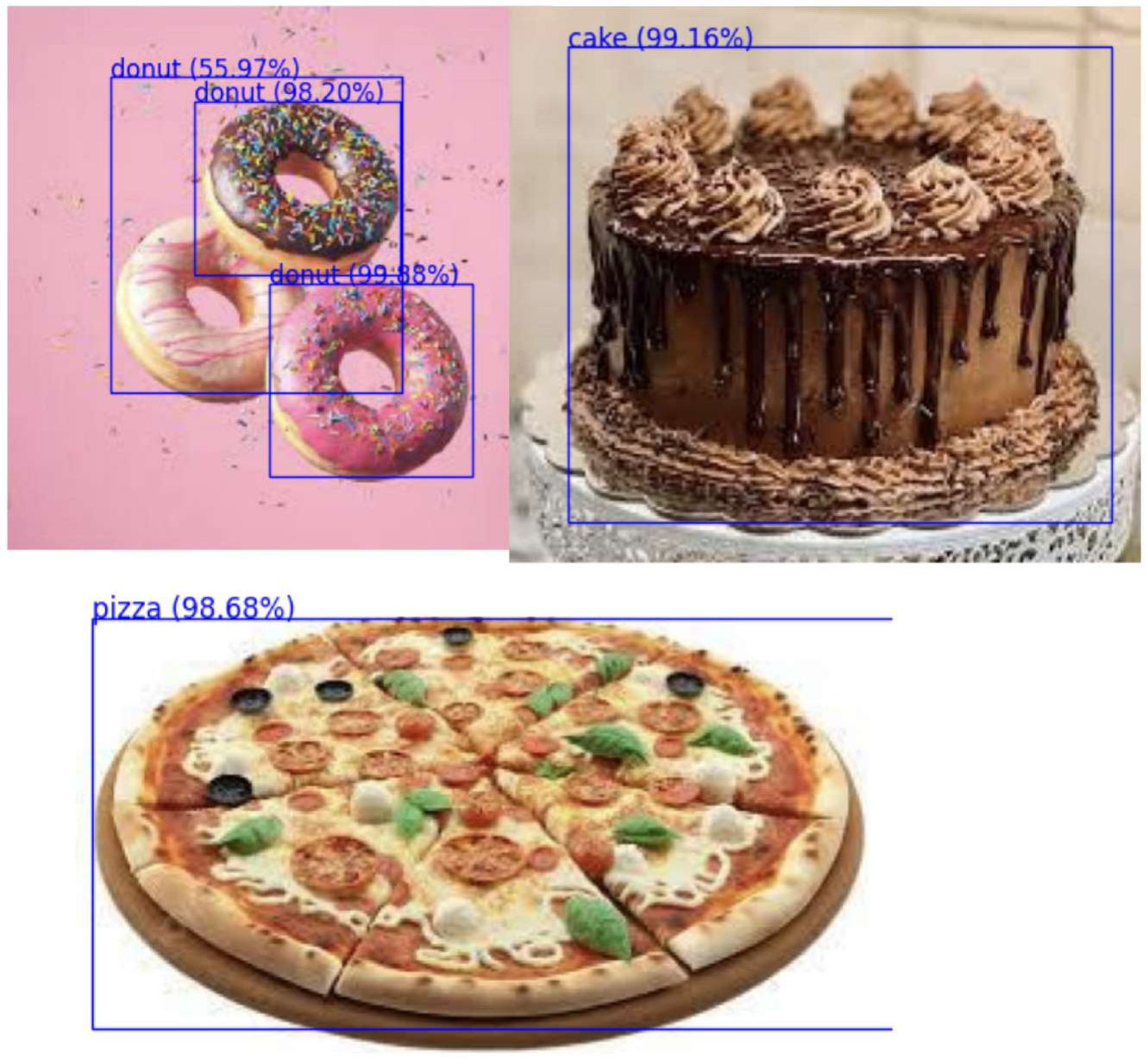}
    \caption{Classification by NutriVision}
    \label{fig:enter-label}
\end{figure}
\\In these situations, the model's confidence scores typically lean toward the middle rather than the high end, suggesting that the model is less certain of its classifications. More diversity in the dataset is considered to be crucial in order to solve these issues and raise the model's confidence scores in complex food plating scenarios. The model would be better able to manage various meal presentations and increase its accuracy under difficult circumstances with a more diverse dataset. Using Algorithm 2 for food quantification on our dataset, Fig. 9 displays the nutritional estimates for two example food plates. This shows that the model can infer nutritional values from the detected and quantified food items in the pictures.
\begin{figure}
    \includegraphics[width=1\linewidth]{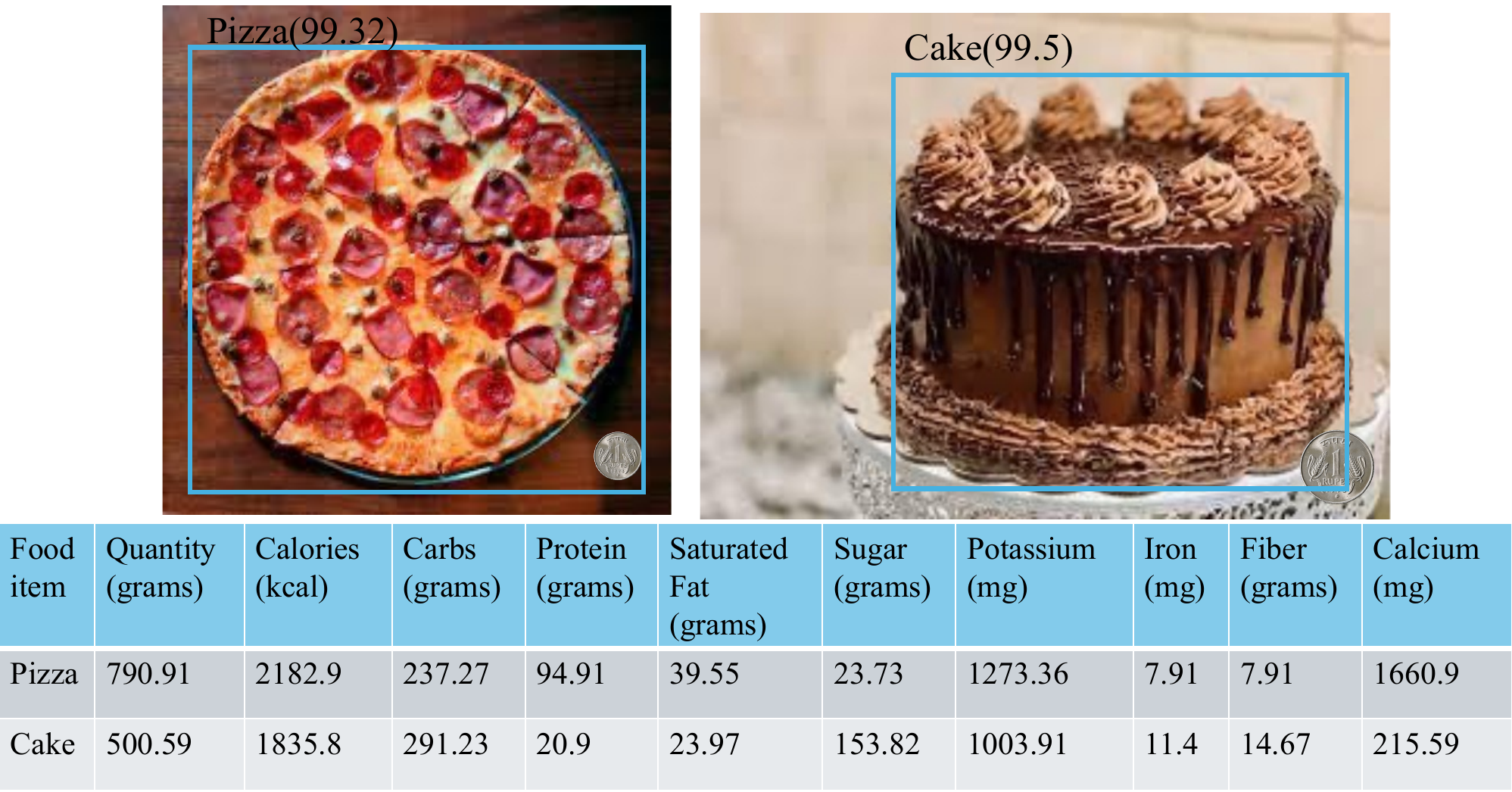}
    \caption{Nutritional Value Estimation by NutriVision}
    \label{fig:enter-label}
\end{figure}
\begin{table}[h]
    \centering
    \caption{Performance Metrics of NutriVision}
    \label{tab:my_label}
    \renewcommand{\arraystretch}{1.5}
    \begin{tabular}{|l|l|l|}
        \hline
        \textbf{Types} & \textbf{Metric} & \textbf{Value} \\
        \hline\hline
        \multirow{3}{*}{Classification (Training)} & Accuracy & 95\% \\ 
        & Precision & 84\% \\
        & Recall & 82\% \\
        \hline
        Localization & IoU & 61\% \\
        \hline
        \multirow{3}{*}{Classification (Testing)} & Accuracy & 92\% \\ 
        & Precision & 82\% \\
        & Recall & 79\% \\
        \hline
    \end{tabular}
\end{table}
\\\\Table 2 provides a comprehensive overview of the NutriVision system's performance data, emphasizing the main functionalities. Several metrics, such as accuracy, precision, recall, and intersection-over-union (IoU), have been calculated for classification purposes, offering a thorough evaluation of the system's performance.
\\\\Detailed graphical representations of the classification loss, localization loss, and overall total loss evolution during the object detector's training phase are shown in Fig. 10. Comprehension of the model's performance dynamics requires a comprehension of these visuals, which provide a thorough grasp of how the object detector changes over time in order to handle classification jobs, achieve localization precision, and control the combined loss metrics.
\\\\The classification loss graph, which decreases during training, illustrates the model's growing accuracy in differentiating between object classes. In the same way, the localization loss graph shows increased accuracy in identifying food items in photos—even in intricate situations. The overall total loss graph shows how well the model locates and recognizes food items by combining the losses from localization and classification by plotting the total loss versus number of steps using Faster RCNN. These visual aids demonstrate NutriVision's present performance as well as opportunities for improvement. The model appears to be becoming more reliable, as evidenced by the gradual decrease in overall loss during training.
\begin{figure}[h]
    \centering
    \includegraphics[width=0.88\linewidth]{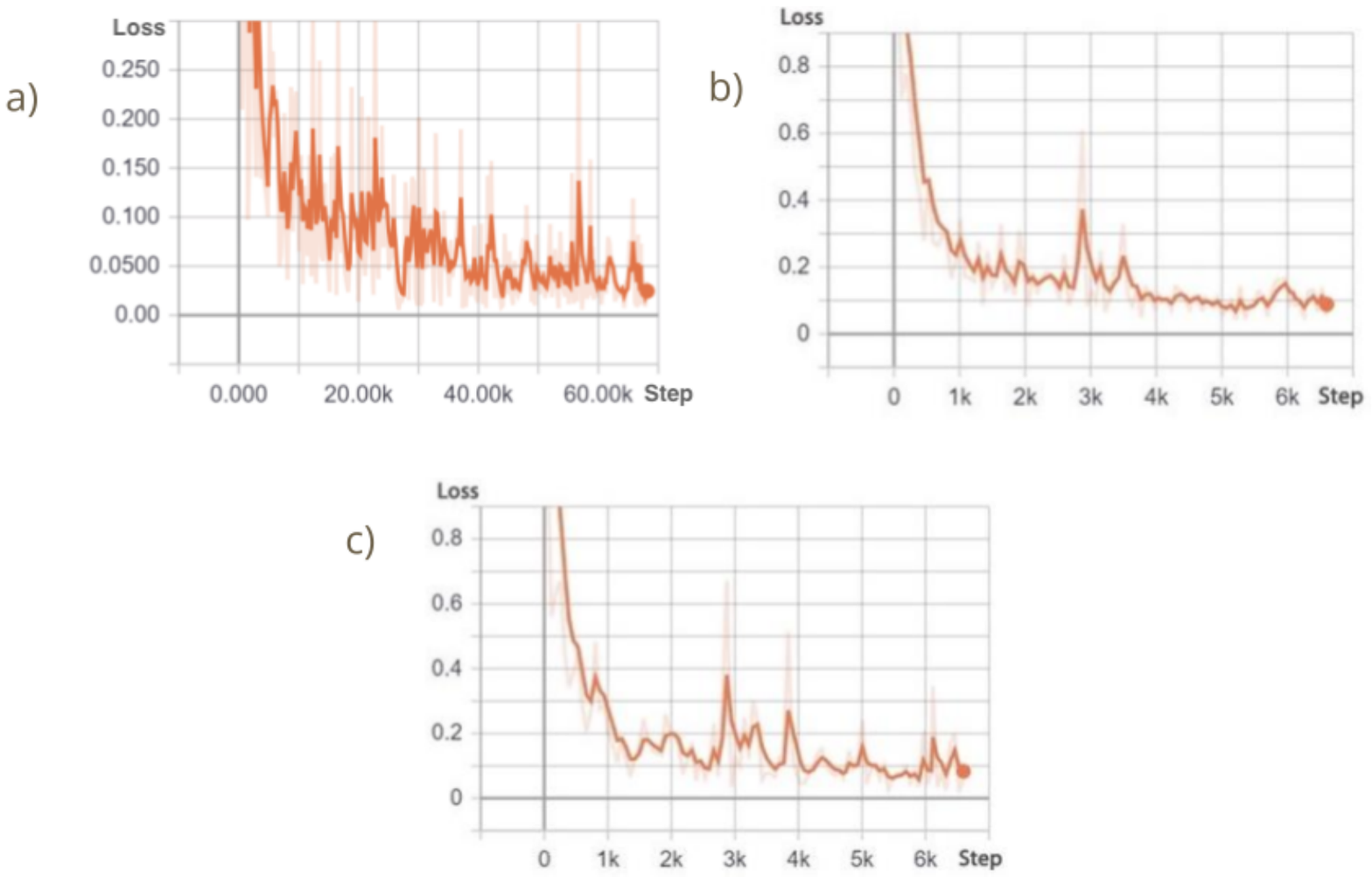}
    \caption{a) Total Loss of Faster RCNN, b) Localization Loss of Faster RCNN, c) Classification Loss of Faster RCNN}
    \label{fig:enter-label}
\end{figure}
\\\\\\The two main visual representations of the data produced by the NutriVision system are the macronutrient distribution graph in Fig. 11 and the macronutrient composition in grams graph in Fig. 12. These graphs, which give both a general summary and a thorough breakdown of the macronutrient composition, are essential for comprehending the nutritional value of the food products that are being studied. 
\begin{figure}[h]
    \centering
    \includegraphics[width=0.55\linewidth]{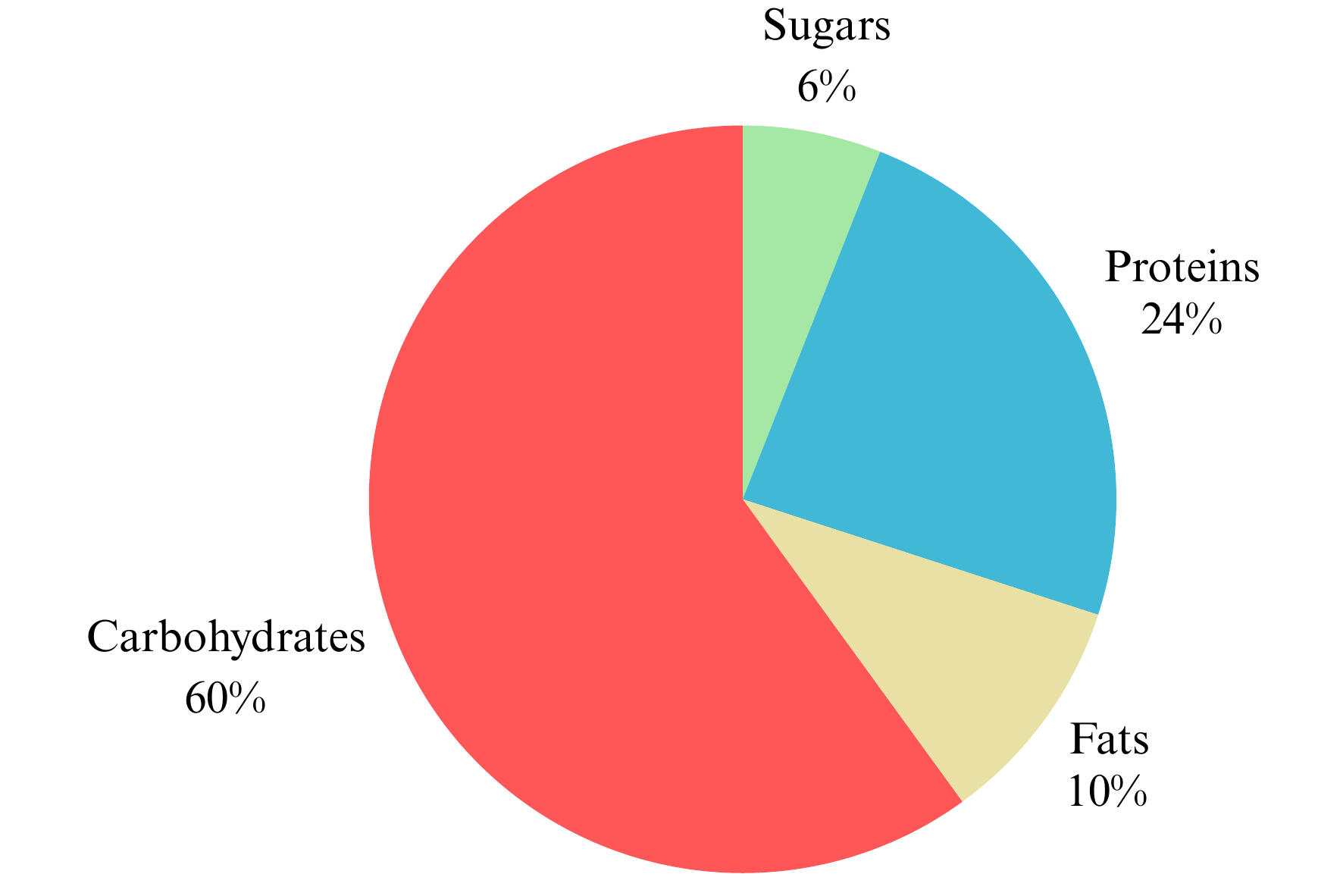}
    \caption{Macronutrient Distribution}
    \label{fig:enter-label}
\end{figure}
\\The macronutrient distribution graph (Fig. 11) breaks down the four main macronutrients (carbohydrates, proteins, sugars, and fats) in the dietary items under analysis as percentages. Users can rapidly ascertain if a meal is strong in carbohydrates, high in protein, or balanced in terms of all macronutrients by referring to this graph, which provides a clear and short assessment of the nutritional balance of a meal. For individuals looking to control their intake of macronutrients, the visual representation is a priceless tool as it provides an instantaneous evaluation of whether a meal fits with their dietary objectives. 
\\On the other hand, the macronutrient composition in grams graph (Fig. 12) gives an accurate assessment of every macronutrient present in the food items. This graph provides precise numbers for the amount of carbs, proteins, carbs, and sugars consumed, measured in grams, in contrast to the percentage-based distribution. Users can precisely trace their consumption by performing nutritional evaluations and dietary planning with this specific composition. This graph helps users make better meal choices by providing information on the precise amounts of each macronutrient, making it easier for them to accurately meet their nutritional needs. When combined, these graphs provide a thorough understanding of meal nutritional composition. Combining these visual aids improves the user's comprehension and control over dietary intake by enabling them to assess food patterns and make educated decisions.
\begin{figure}[h]
    \centering
    \includegraphics[width=0.85\linewidth]{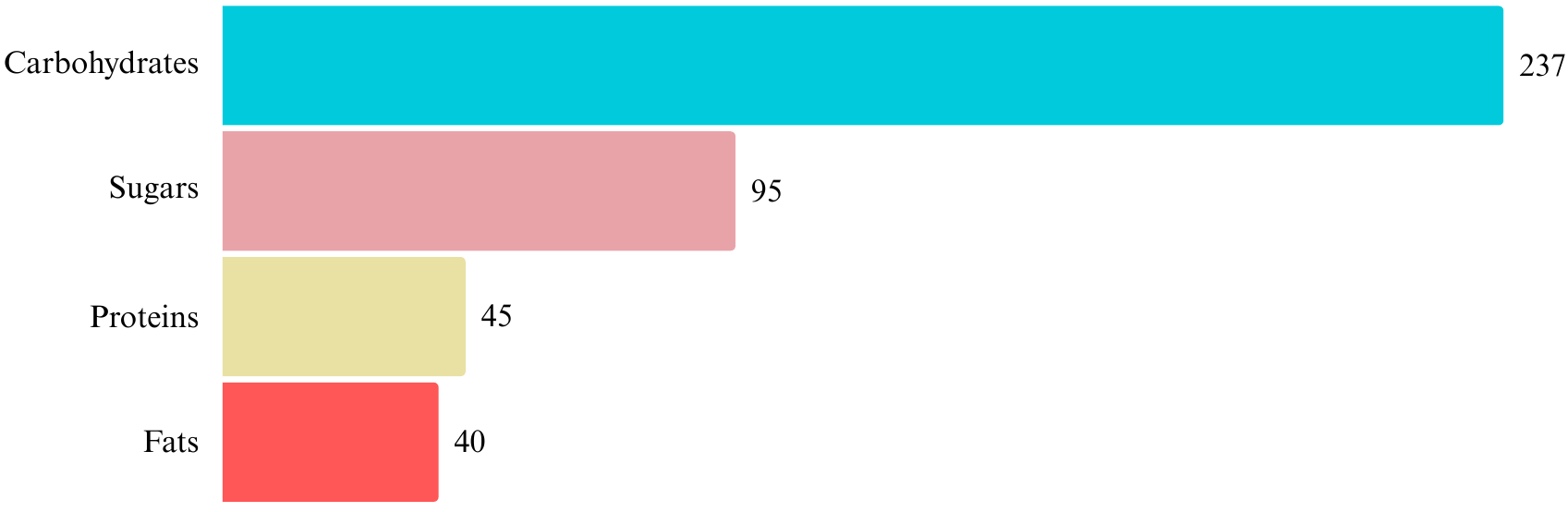}
    \caption{Macronutrient Composition in grams}
    \label{fig:enter-label}
\end{figure}
\begin{figure}[h]
    \centering
    \includegraphics[width=0.58\linewidth]{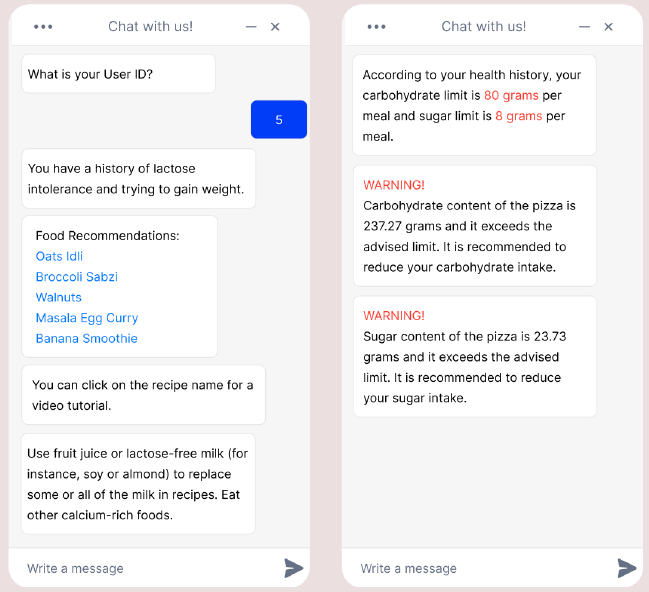}
    \caption{Personalized Diet Advice}
    \label{fig:enter-label}
\end{figure}
\\\\Fig. 13 presents an intuitive user interface that walks users through a customized dietary analysis procedure, greatly improving the user experience. Users are required to enter their unique user ID, which is connected to their dietary and personal health information, when they first use the interface. After the user ID is entered, the system processes the data using sophisticated algorithms to provide the user with individualized nutritional recommendations based on their health history. Next, based on the user's individual dietary requirements, the interface shows the top five food recipes that are suggested. Based on the user's dietary preferences, nutritional objectives, and medical history, these recommendations have been carefully selected. Users can easily follow along and replicate the dishes at home with the help of the interface, which also offers a link to a corresponding recipe video on YouTube.
\\Furthermore, an essential component of the interface raises user awareness of possible nutritional concerns. A text alert alerts the user if a suggested meal has more sugar or carbohydrates than is healthy for them. Because each user has different limits on how much sugar and carbohydrates they can consume based on their dietary preferences and medical history, this alarm system is tailored specifically for them. With customized restrictions kept in a separate database connected to user IDs, these notifications are customized for each individual user, guaranteeing feedback and monitoring that is specific to them.
\\\\Through the integration of individualized alerts, the NutriVision system guarantees users are aware of any possible nutritional issues, empowering them to make healthier and safer choices. Overall, the NutriVision system's dedication to offering a thorough, customized, and user-friendly nutritional analysis tool that enables users to make educated and health-focused dietary decisions is highlighted by the combination of these visual representations and the intuitive interface in Fig. 13.
\begin{figure}[h]
    \centering
    \includegraphics[width=0.3\linewidth]{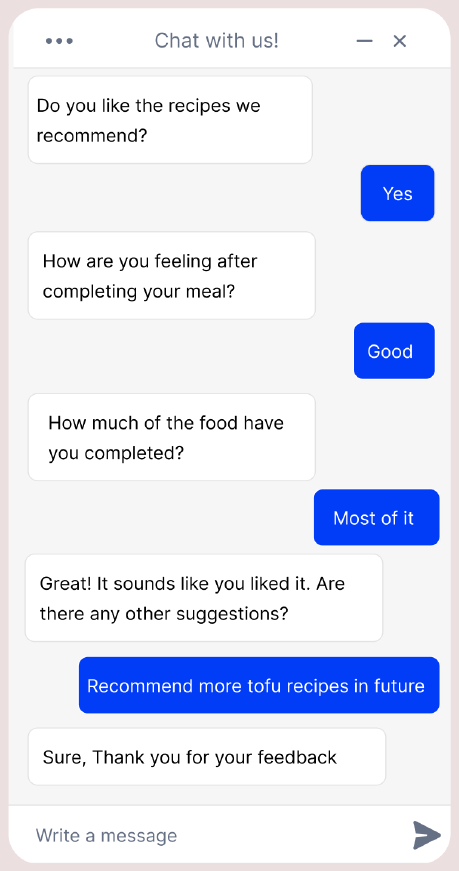}
    \caption{Feedback System of NutriVision}
    \label{fig:enter-label}
\end{figure}
\\\\A simple feedback exchange is shown in Fig. 14, where the user is asked about their experience using the recommended recipes. This exchange facilitates the collection of basic information regarding user participation and meal completion. It also offers insightful information to enhance the recommendation system. By analyzing user feedback, the system can adapt to individual preferences and improve the quality of future recommendations. This iterative process not only increases user satisfaction but also fosters a more personalized and engaging culinary experience.
\\\\The utilization of input data, such as height, weight, and gender, to calculate a user's Body Mass Index (BMI) is another feature of NutriVision. Following calculation, this BMI is divided into four groups: underweight, normal weight, overweight, and obese. NutriVision creates customized meal suggestions based on a user's BMI category, taking into account their dietary preferences, whether they are vegan, non-vegetarian, or vegetarian. The suggestion algorithm of the system is made to guarantee a diet rich in variety. 
\\\\In addition to taking into account the user's BMI and dietary preferences, it assesses the nutritional value of meals the user has previously eaten. NutriVision can detect any potential nutritional gaps, such as an inadequate intake of carbohydrates, proteins, or fats, by examining the nutritional values of these previous meals. Subsequently, the algorithm customizes its meal recommendations to compensate for these inadequacies, encouraging a well-rounded diet that enhances the user's general health and wellbeing.
\begin{figure}[h]
    \centering
    \includegraphics[width=0.87\linewidth]{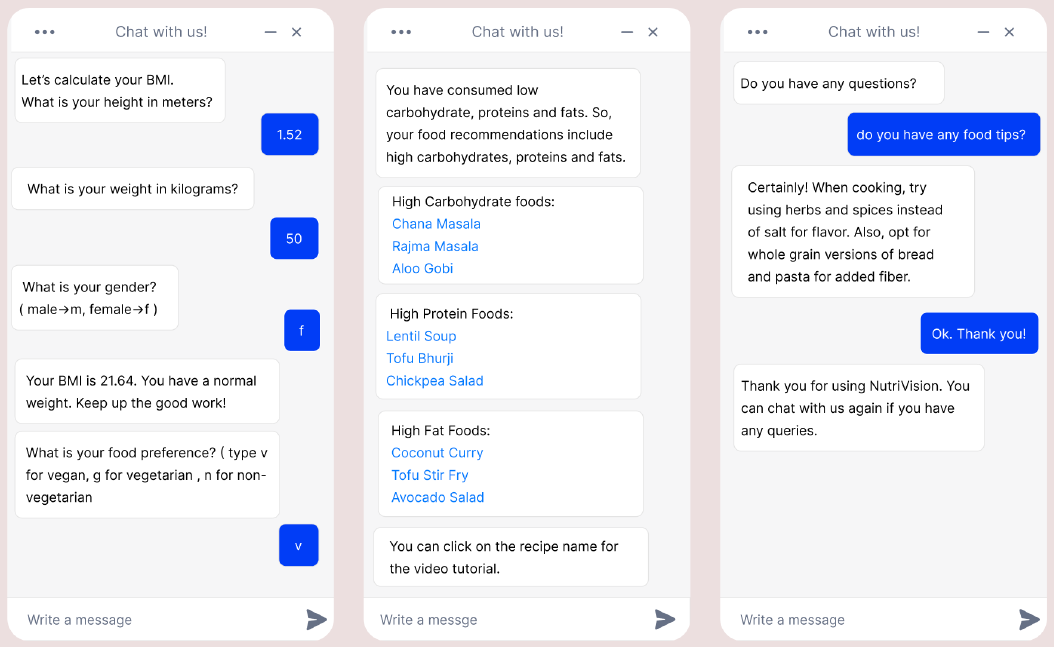}
    \caption{Food Recommendation System-Case(i)}
    \label{fig:enter-label}
\end{figure}
\\\\Fig. 15 presents a detailed interaction between the chatbot and a female user who maintains an average weight and strictly follows a vegan diet. This figure highlights how NutriVision adeptly customizes its meal recommendations to cater to the user's particular dietary preferences and overall health condition. By doing so, NutriVision ensures that the recommendations provided are not only aligned with her specific needs but also promote a well-balanced and nutritious diet, demonstrating its capability to adapt to various dietary restrictions while maintaining nutritional integrity.
\begin{figure}[h]
    \centering
    \includegraphics[width=0.87\linewidth]{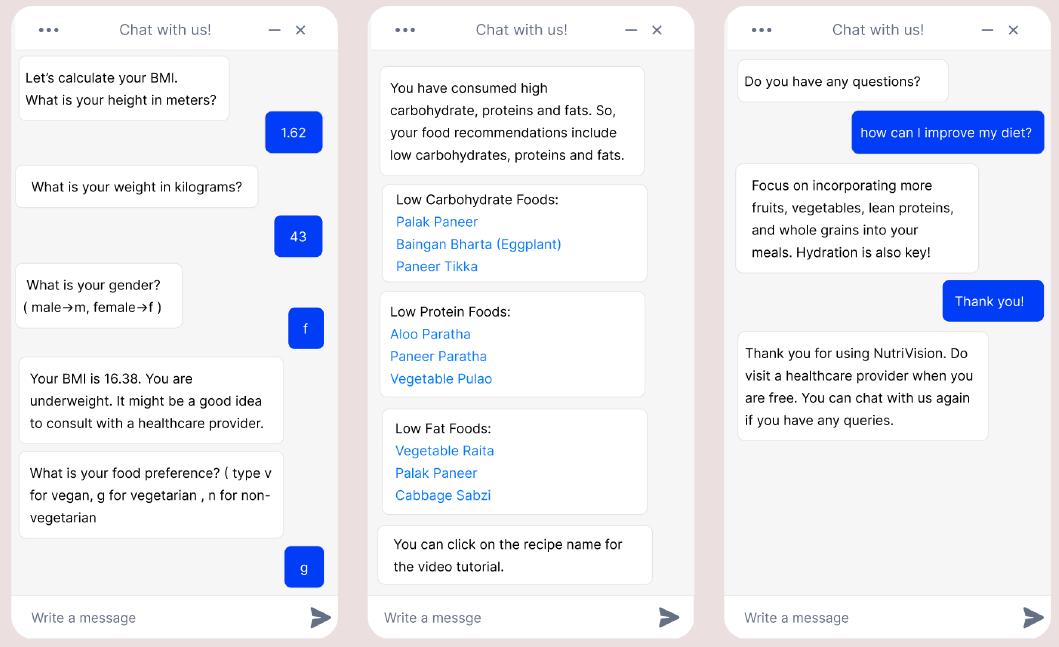}
    \caption{Food Recommendation System-Case(ii)}
    \label{fig:enter-label}
\end{figure}
\\\\On the other hand, Fig. 16 depicts NutriVision's tailored suggestions for a female user classified as underweight. This figure illustrates the system's capability to adjust its recommendations based on the user's specific BMI category. By offering carefully considered nutritional guidance, NutriVision focuses on helping the user achieve a healthier weight through a strategic selection of foods. The system's ability to adapt its recommendations according to the user's unique health profile underscores its effectiveness in providing personalized dietary advice aimed at promoting overall well-being. This personalized approach not only enhances user satisfaction but also empowers individuals to make informed choices that support their health goals.
\begin{figure}[h]
    \centering
    \includegraphics[width=0.87\linewidth]{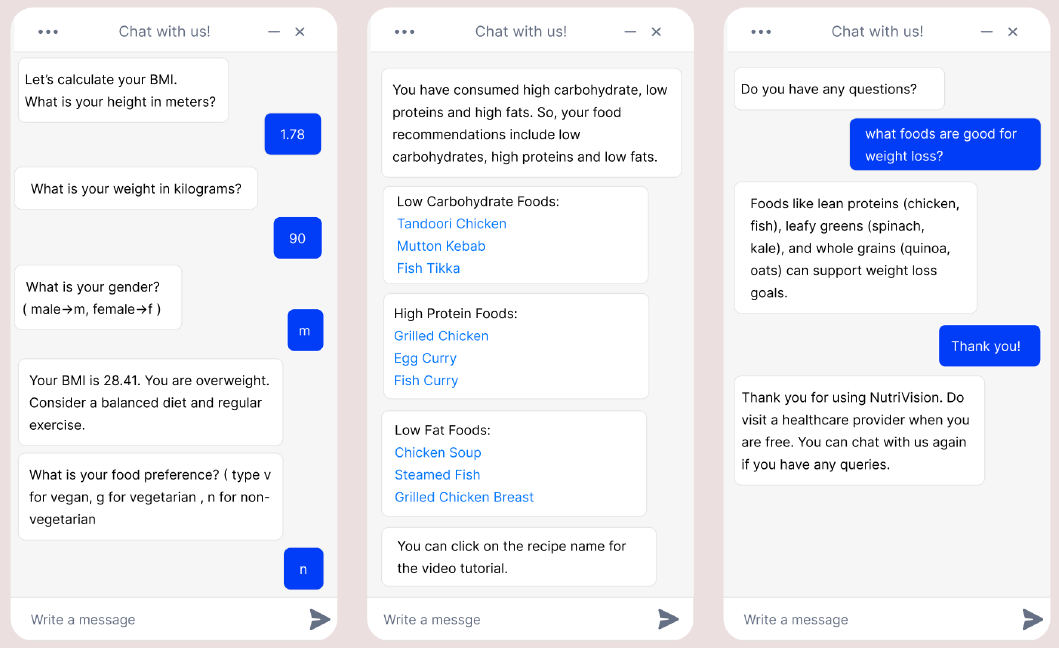}
    \caption{Food Recommendation System-Case(iii)}
    \label{fig:enter-label}
\end{figure}
\\\\Fig. 17 illustrates the dietary recommendations provided by NutriVision for an overweight male user, further emphasizing the system’s remarkable versatility. NutriVision's method is designed to deliver personalized meal recommendations that actively support the user in achieving a healthier weight, considering the distinct dietary needs and weight management goals associated with this particular BMI category. Beyond its ability to generate tailored meal plans, NutriVision also boasts an interactive chatbot function that significantly enhances user engagement and customer support. Users have the flexibility to engage with the chatbot by asking detailed questions about nutrition, health-related topics, or specific meal recommendations provided by the system. This interactive feature ensures that users can receive immediate guidance and clarifications, transforming NutriVision into a dynamic resource for ongoing nutritional education and personalized meal planning.
\\\\Moreover, NutriVision’s commitment to fostering optimal health through individualized, data-driven nutrition guidance is vividly demonstrated by its holistic approach, which integrates BMI-based dietary recommendations with its interactive chatbot feature. The system emerges as a highly effective tool for assisting users in achieving and maintaining a balanced and nutritious diet, thanks to its ability to adapt recommendations based on a comprehensive understanding of the user's BMI, dietary preferences, and previous nutritional intake. This makes NutriVision not only a practical solution for personalized meal planning but also a valuable educational resource that empowers users to make informed decisions about their nutrition and overall health.

\subsection{Comparison with existing research}
While the system in \cite{b1} can identify meals and estimate calories, it does not provide personalized advice or comprehensive calorie estimations that are customized for each user. In a similar way, \cite{b2} provides identification tools along with nutritional information on a variety of cuisines, although it does not offer tailored dietary suggestions. The literature study reveals a clear gap in current systems: some concentrate on food recognition, but they frequently fail to automate quantity estimation, requiring food amounts to be manually entered. 
\\NutriVision, on the other hand, stands out as a unique remedy that successfully resolves these shortcomings. It outperforms systems such as \cite{b1} and \cite{b2} in that it incorporates extensive functionalities that cover meal recognition and calorie estimation in addition to offering tailored advice and accurate nutritional assessments. NutriVision stands out from its predecessors by providing a comprehensive and personalized nutritional assessment. It markets itself as a comprehensive and cutting-edge solution for consumers looking for precise and tailored dietary guidance.
\\\\Compared with \cite{b1} and \cite{b2} object identification techniques, NutriVision uses Faster R-CNN, a region-based convolutional neural network (CNN) adaption that is well-known for its higher efficacy in picture classification and localization tasks. When compared to conventional CNN techniques, this sophisticated adaption of CNN significantly enhances image processing speed and efficiency. In numerous crucial areas related to nutritional calculation and user guiding, faster R-CNN as used in NutriVision displays distinct advantages over traditional CNN algorithms. It is particularly good at differentiating between things that look alike, which is important when giving proper dietary recommendations for foods that might look alike. This feature is especially helpful in guaranteeing accurate dietary advice. 
\begin{table}[h]
    \centering
    \caption{Comparison of Object Detection Algorithms}
    \label{tab:my_label}
    \setlength{\tabcolsep}{12pt} 
    \renewcommand{\arraystretch}{1.5} 
    \begin{tabular}{|c|c|c|c|c|}
        \hline
        Metric & Faster R-CNN & CNN & Mask R-CNN & YOLO \\
        \hline\hline
        Accuracy & 90\% & 84\% & 87\% & 82\% \\
        \hline
        Precision & 81\% & 77\% & 79\% & 75\% \\
        \hline
        Recall & 79\% & 73\% & 76\% & 70\% \\
        \hline
    \end{tabular}
\end{table}
\\\\Additionally, NutriVision makes use of Faster R-CNN's ability to identify several food items on a single plate—a crucial component of a thorough nutritional analysis. Faster R-CNN improves accuracy and dependability in nutritional analysis applications by precisely identifying and differentiating minute visual differences between different food items thanks to its fine-grained detection capabilities. The performance metrics for the NutriVision dataset shown in Table 3 illustrate how Faster R-CNN's stronger localization ability translates into higher precision and accuracy in food detection.
\\\\Our experimental results show that, over a wide variety of assessment parameters, Faster R-CNN performs consistently better than YOLO, Mask R-CNN, and conventional CNN models. The Evaluation Metrics graphs for several object detection algorithms given in Fig. 18 clearly demonstrate this higher performance.
\\\\The Training Time vs. Epochs graph (Fig. 18a) demonstrates that Faster R-CNN exhibits greater accuracy and convergence while requiring a longer training period. This graph illustrates how Faster R-CNN outperforms the other models in terms of increasing accuracy over time. Faster R-CNN yields the highest values in terms of F1 score, indicating a better trade-off between recall and precision. Fig. 18b shows the F1-Score vs. Epochs graph, which reveals that Faster R-CNN consistently achieves a higher F1 score.
\\\\Furthermore, the graphs displaying Recall vs. Epochs (Fig. 18c) and Precision vs. Epochs (Fig. 18d) support the superiority of Faster R-CNN. These graphs highlight the efficacy of Faster R-CNN in reliably detecting objects by showing that it consistently achieves the highest recall and precision values. These thorough results show that Faster R-CNN is the most trustworthy and efficient method for achieving overall detection accuracy. It is hence the best option for tasks involving object detection.
\\\\The most distinctive characteristic of NutriVision that other modern platforms do not offer is the degree of customisation offered. Personalized dietary suggestions are made by NutriVision using a thorough examination of the user's past and BMI. The system can provide the user with tailored recommendations that promote long-term health goals while also satisfying their immediate nutritional demands by incorporating historical food decisions, health conditions, and BMI data. 
\\\\NutriVision is different from other systems because of its advanced user feedback and customization features, which allow for a highly customized and dynamic user experience. Users have the ability to directly comment on the meal recommendations they get, regardless of whether they are based on dietary requirements, taste preferences, or particular nutritional objectives. Through this feedback system, NutriVision is able to continuously improve and modify its recommendations so that they more closely suit the unique requirements and preferences of each user. Through this iterative process, the system's recommendations are continuously refined, leading to a more precise and fulfilling nutritional plan. 
\begin{figure}[h]
\centering
\begin{subfigure}[b]{0.48\textwidth}
    \centering
    \includegraphics[width=\linewidth, height=6cm]{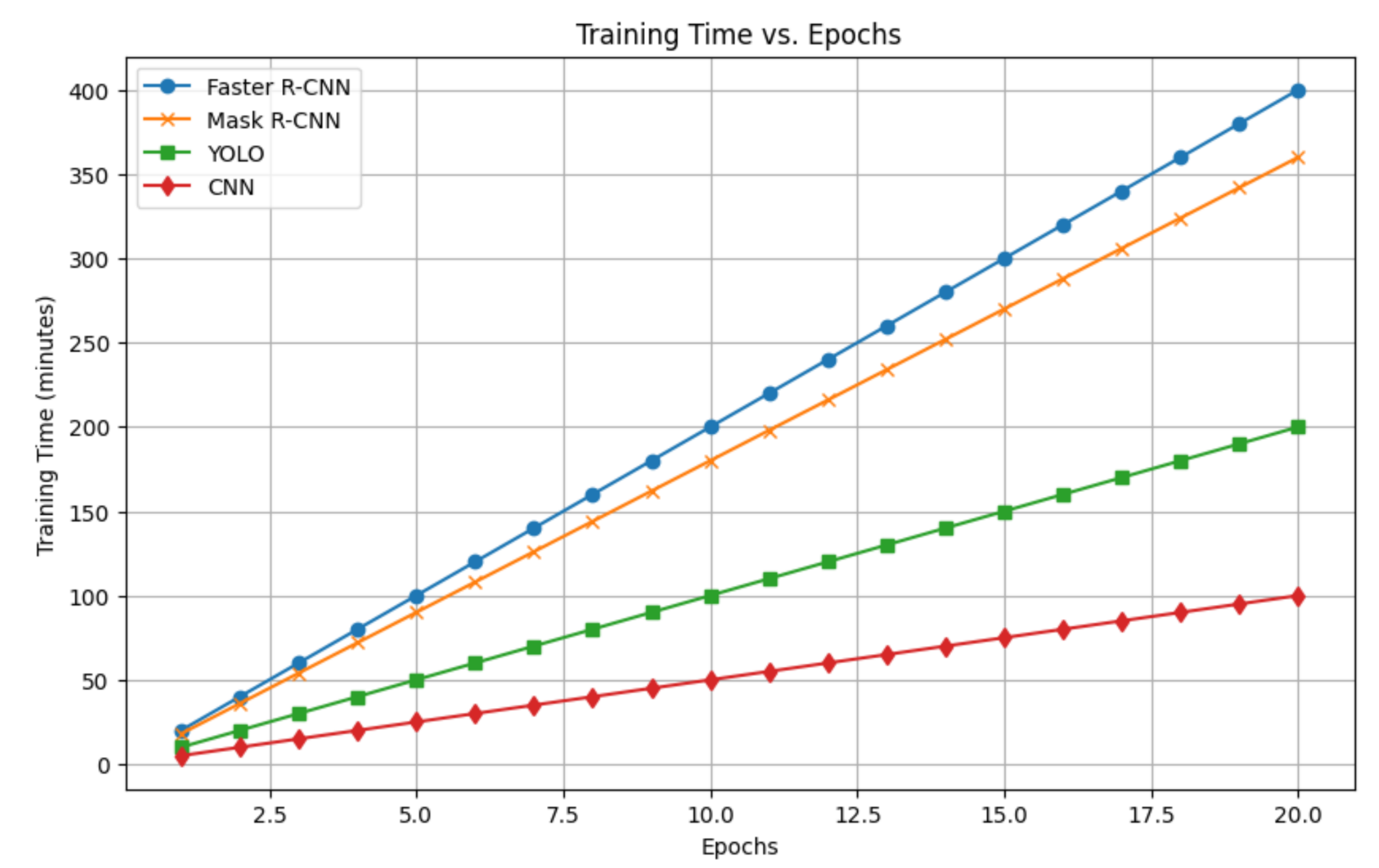}
    \caption{Training Time vs Epochs Graph}
    \label{fig:subim1}
\end{subfigure}
\hfill
\begin{subfigure}[b]{0.48\textwidth}
    \centering
    \includegraphics[width=\linewidth, height=6cm]{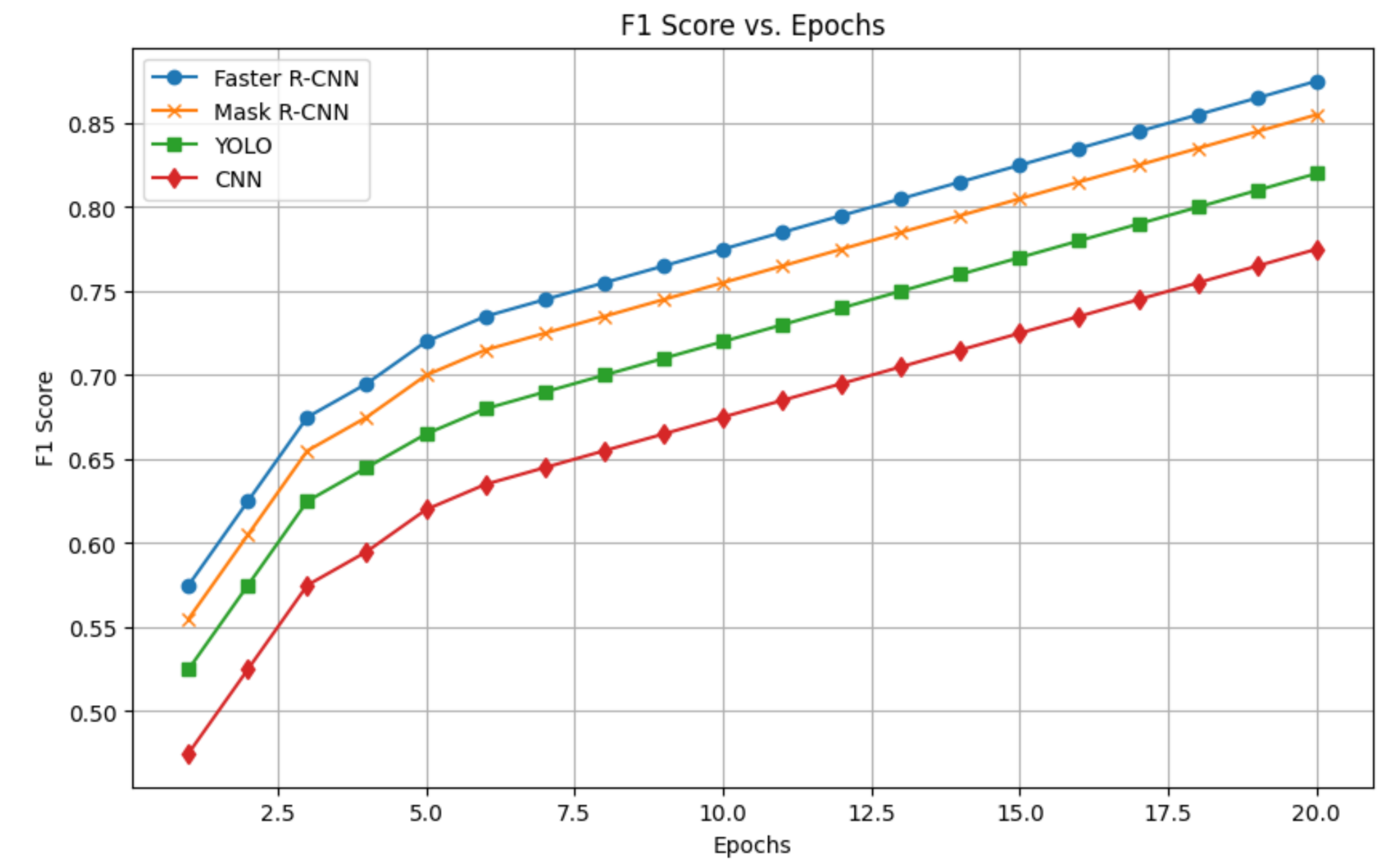}
    \caption{F1-Score vs Epochs Graph}
    \label{fig:subim2}
\end{subfigure}
\vskip\baselineskip
\begin{subfigure}[b]{0.48\textwidth}
    \centering
    \includegraphics[width=\linewidth, height=6cm]{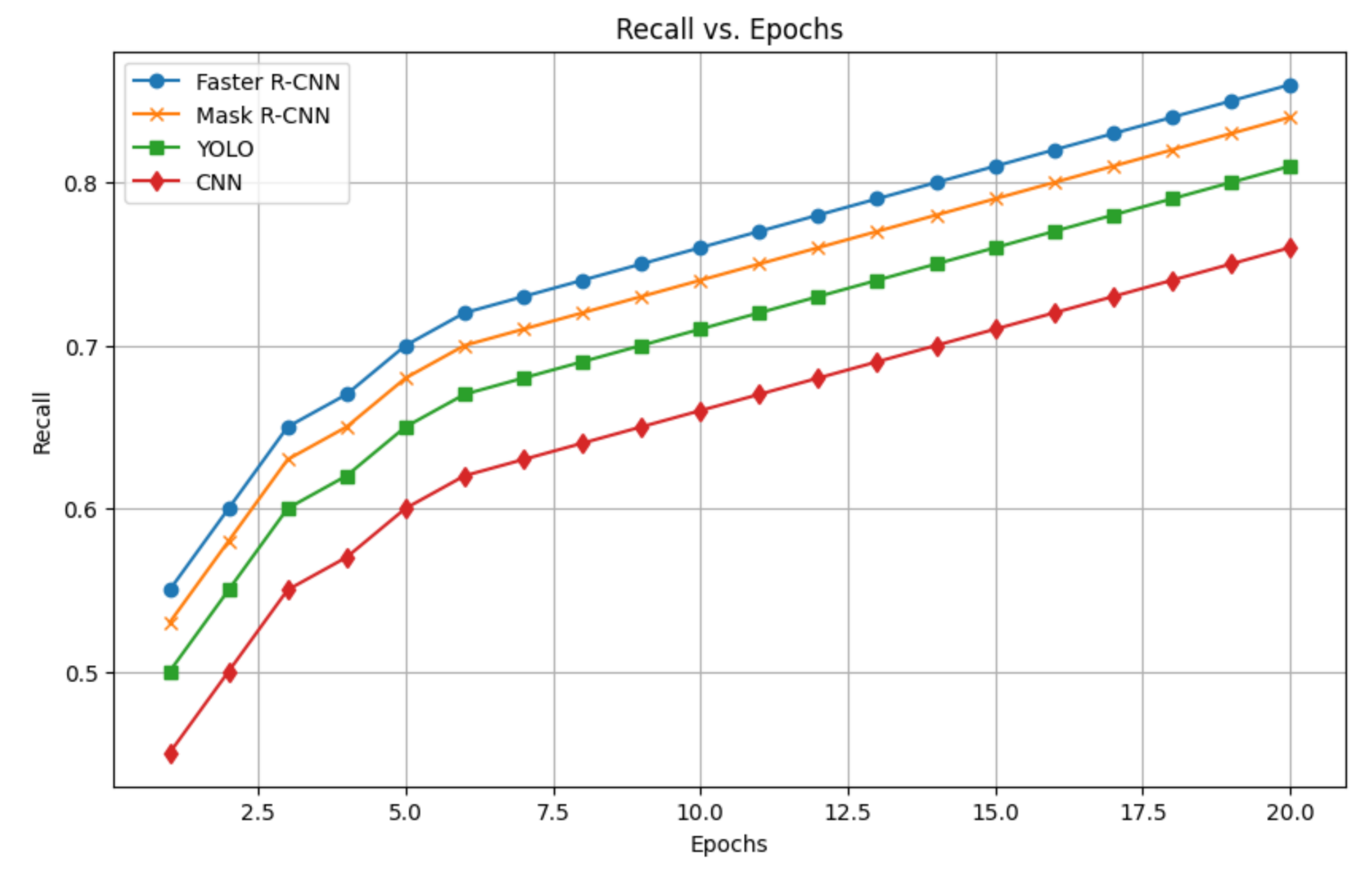}
    \caption{Recall vs Epochs Graph}
    \label{fig:subim3}
\end{subfigure}
\hfill
\begin{subfigure}[b]{0.48\textwidth}
    \centering
    \includegraphics[width=\linewidth, height=6cm]{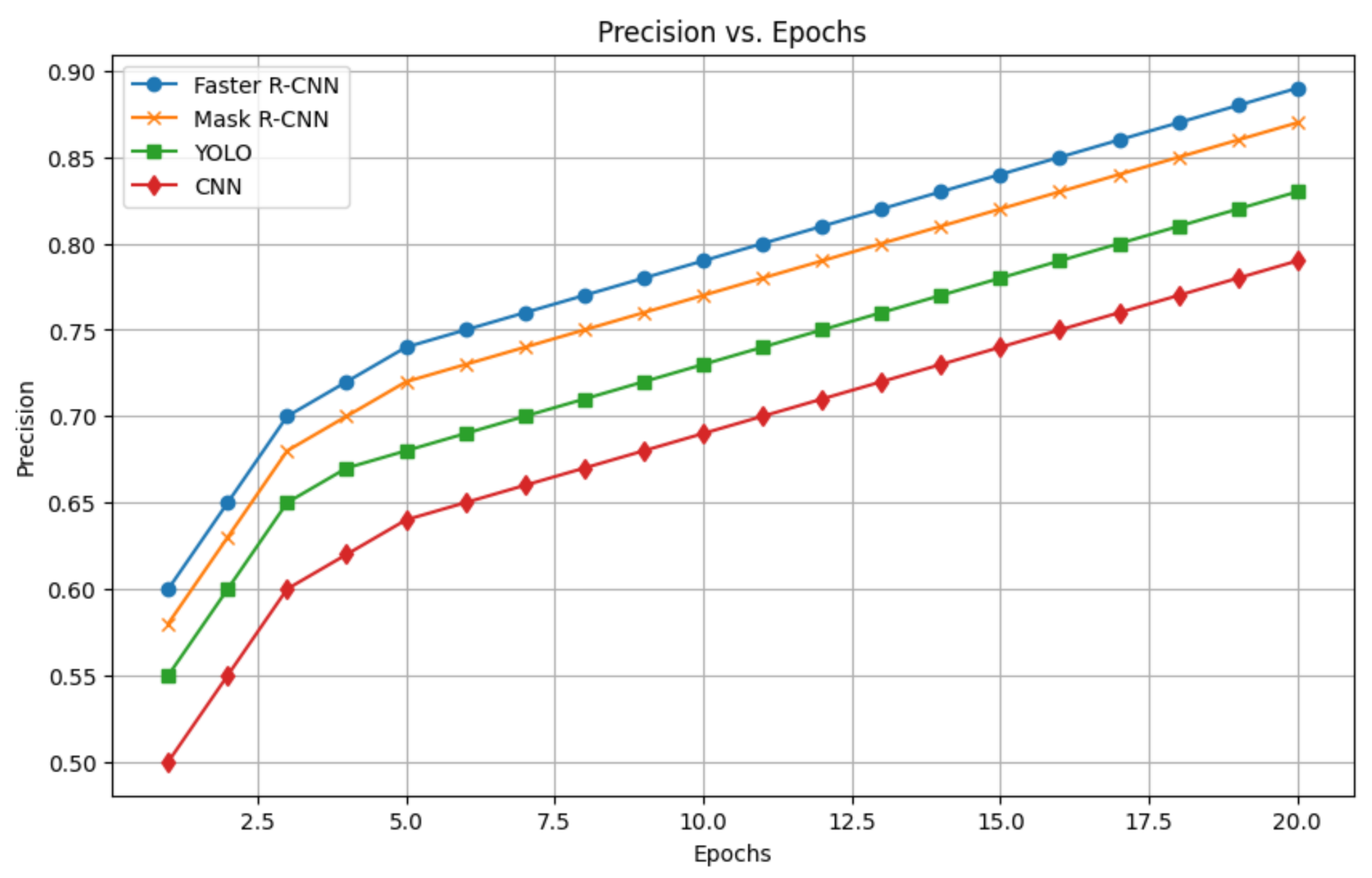}
    \caption{Precision vs Epochs Graph}
    \label{fig:subim4}
\end{subfigure}
\caption{Evaluation metrics graphs of different object detection algorithms}
\label{fig:image2}
\end{figure}
\section{Conclusion and Future Work}  
\label{sec:conclusion}
When it comes to our health, food is paramount. Studies \cite{b3}, \cite{b4} show that preventing diseases largely depends on maintaining a healthy diet, making it crucial to monitor not just calorie intake but also other nutrients. This paper proposes the NutriVision system for food detection, nutritional value estimation, and personalized diet advice. NutriVision uses advanced algorithms like Faster R-CNN for accurate meal identification, enhancing food detection and nutritional analysis reliability. The system automates nutritional value calculation from food images, improving accuracy by determining food volume and converting it to weight. However, it has a limitation in accounting for partially consumed meals. Personalized recommendations are based on the user's health profile, including BMI and dietary preferences, ensuring tailored suggestions. The user-friendly interface allows users to input their unique ID and receive customized dietary advice, including recipe recommendations and links to instructional videos, improving the overall user experience. The system offers real-time alerts for excessive sugar or carbohydrate intake, helping users make informed decisions, and continuously learns from user feedback to refine its recommendations.
\\\\In our forthcoming endeavors, we aim to introduce a food quantification system that does not use any reference, emphasizing the need to improve precision and Intersection over Union (IoU) metrics. This system will undergo expansion to accommodate a broader array of food items, enhancing its versatility and applicability. Additionally, each cuisine category within the system is slated to receive updates. Future proposals encompass implementing cutting-edge precision techniques, leveraging the latest in machine learning and computer vision advancements, designing a user-friendly interface, enabling real-time updates for adaptability, and integrating the system with dietary guidelines to offer users insightful nutritional information for health-conscious choices.

\begin{small}

\end{small}
\vspace{0cm}
\begin{minipage}{0.25\textwidth}
  \includegraphics[height=1.5in]{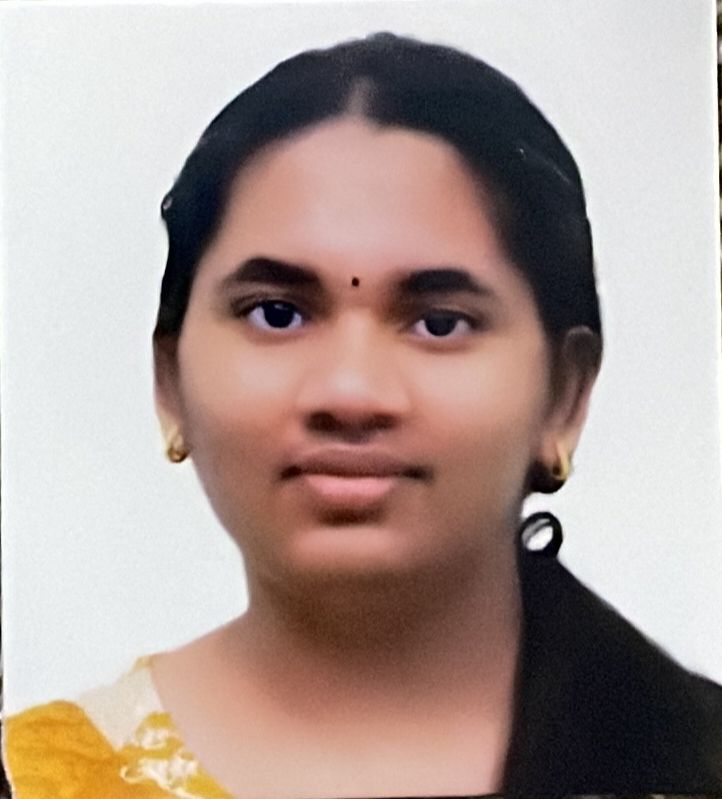} 
\end{minipage}%
\begin{minipage}{0.8\textwidth}
  \textbf{Madhumita Veeramreddy} is a Bachelor of Technology student at SRM University, Amaravati (SRMAP). She specializes in Artificial Intelligence and Machine Learning. Her research interests include developing advanced algorithms for image recognition and classification, optimizing machine learning models for real-time applications, and exploring the intersection of AI with healthcare and nutrition. 
\end{minipage}

\vspace{1cm}
\begin{minipage}{0.25\textwidth}
  \includegraphics[height=1.5in]{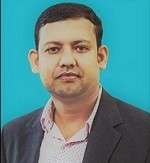} 
\end{minipage}%
\begin{minipage}{0.8\textwidth}
  \textbf{Ashok Kumar Pradhan} is currently working as an Associate Professor in the Department of Computer Science \& Engineering, School of Engineering and Applied Science at SRM University, Amaravati, AP. He earned his M.Tech degree in Computer Science and Engineering from the National Institute of Technology (NIT), Rourkela in 2010, and completed his Ph.D. at NIT Durgapur in 2015. His research interests span across several cutting-edge domains including Optical Communication and Networks, the Internet of Things (IoT), Blockchain Technology, Cyber Security \& Privacy, Machine Learning (ML) \& Deep Learning (DL), Cloud Computing, Edge Computing, Fog Computing, and Computer Algorithms. He has published over 35 research papers in reputed peer-reviewed journals and conferences, edited two books, and contributed four book chapters published by leading academic publishers. He has also been granted one patent and successfully supervised one PhD scholar to completion. In 2019, he was awarded the prestigious SERB grant (TAR/2019/000286). In addition to his research contributions, he serves as a reviewer for renowned journals and transactions published by Springer, Elsevier, and IEEE, helping to ensure the quality and impact of research in his fields of expertise. 
\end{minipage}

\vspace{1cm}
\begin{minipage}{0.25\textwidth}
  \includegraphics[height=1.5in]{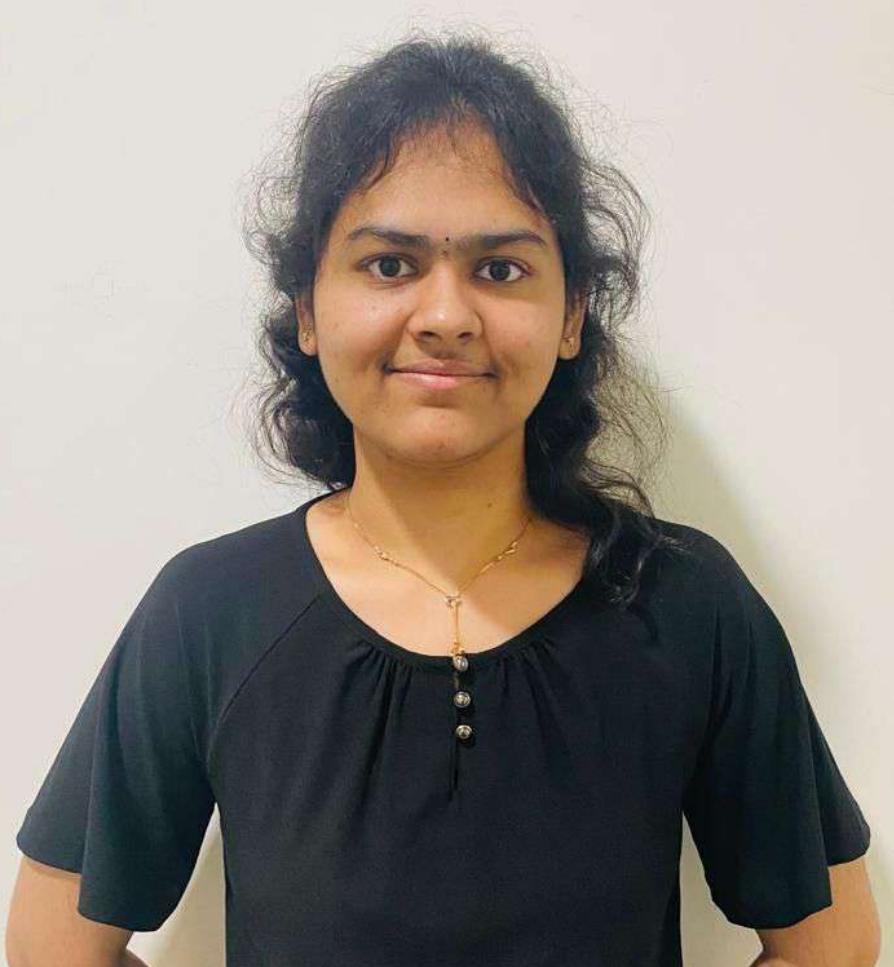} 
\end{minipage}%
\begin{minipage}{0.8\textwidth}
\textbf{Swetha Ghanta} received her Bachelor of Technology (B. Tech.) degree and Master of Technology (M. Tech.) degree in Computer Science and Engineering from RVR \& JC College of Engineering, Guntur, India. She is currently pursuing her PhD degree in the Department of Computer Science and Engineering at SRM University, Amaravati, India. Her research interests include deep learning, enhanced privacy, and security approaches. She is also working on medical image analysis using Federated Learning.
\end{minipage}
\vspace{1cm}
\begin{wrapfigure}{l}{0.18\textwidth}
    \includegraphics[height=1.5in]{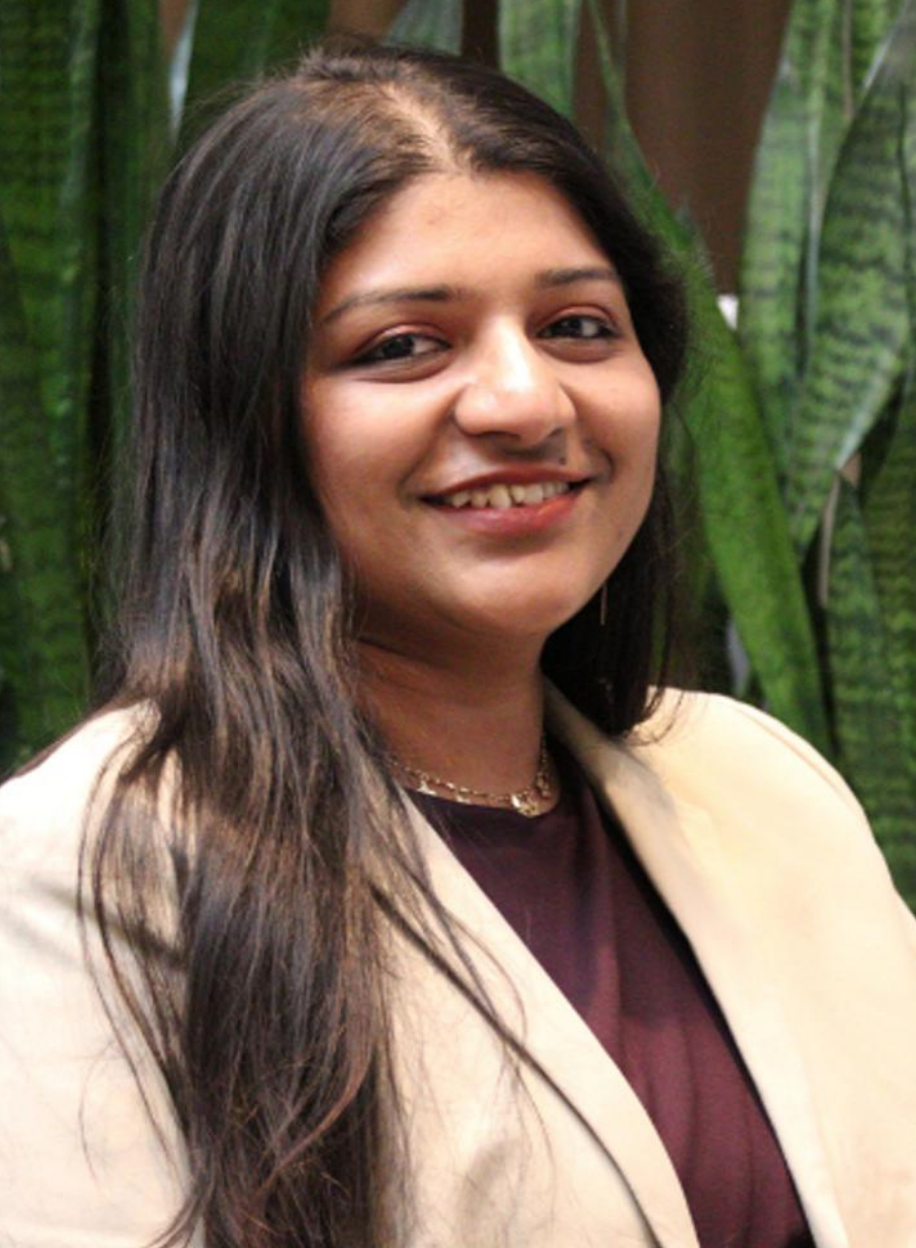}
\end{wrapfigure}
\textbf{Laavanya Rachakonda} (M’21) is an Assistant Professor in the Department of Computer Science in the College of Science and Engineering at the University of North Carolina Wilmington, Wilmington, NC. She earned her Bachelor of Technology (B. Tech) in Electronics and Communication from Jawaharlal Nehru Technological University (JNTU), Hyderabad, India, Master of Sciences (M.S) in Computer Engineering, and Doctor of Philosophy (Ph.D.) in Computer Science and Engineering from University of North Texas. During her graduate studies, she was part of the Smart Electronics Systems Laboratory (SESL) research group at Computer Science and Engineering at the University of North Texas, Denton, TX. Her research interests include smart healthcare applications using artificial intelligence, deep learning approaches, and application-specific architectures for consumer electronic systems based on the IoT. She has 3 peer-reviewed journals published, 13 peer-reviewed conference publications, 2 filed patents, and 1 patent disclosure. She has delivered 15 talks (online and offline) at various IEEE-hosted conferences. She has won 20 honors and awards and has monitored 6 undergraduate and TAMS students. Her biography, research, education, and outreach activities are available at \url{www.laavanyarachakonda.com} 
\vspace{1cm}

\begin{wrapfigure}{l}{0.18\textwidth}
    \includegraphics[height=1.5in]{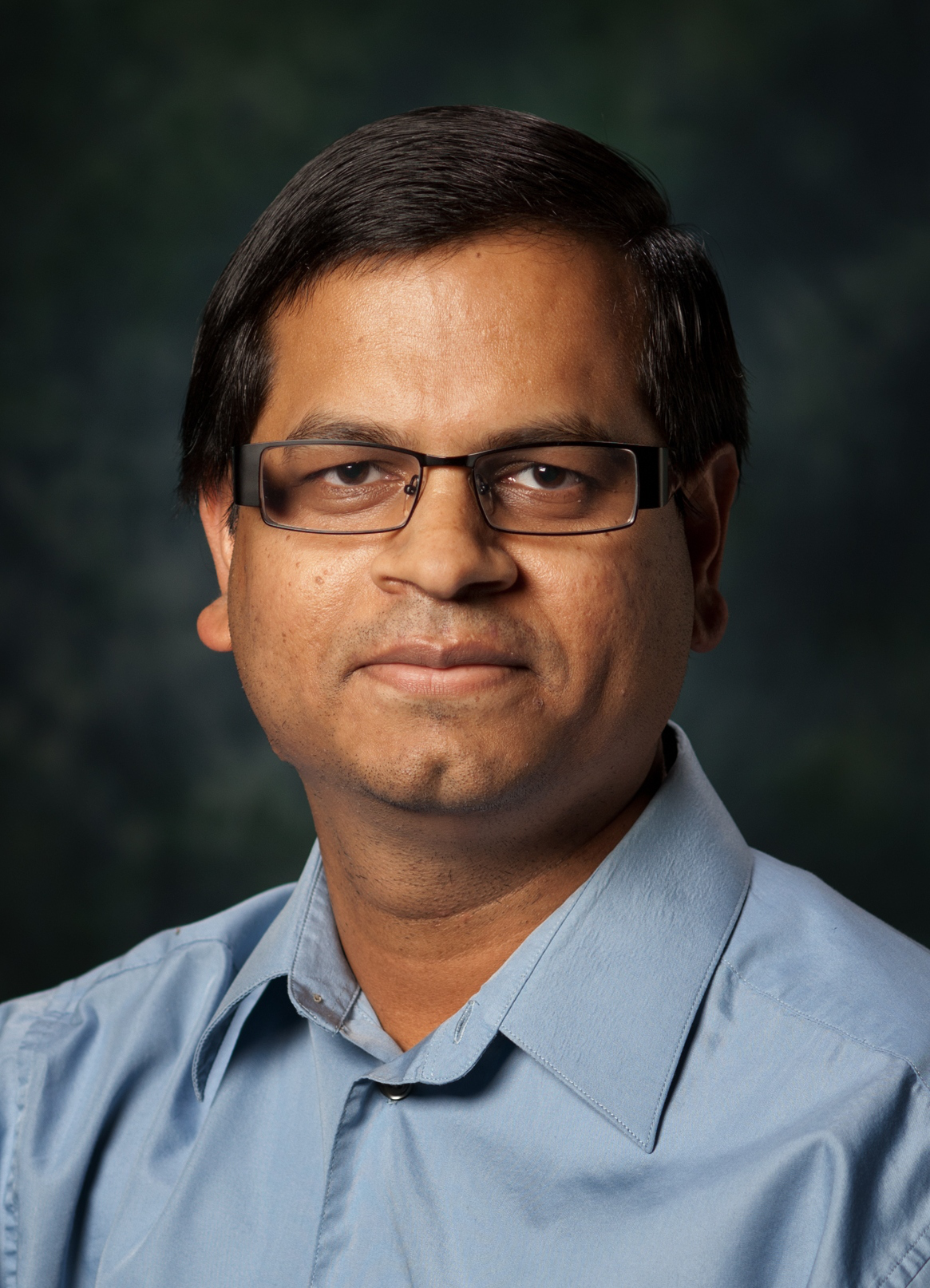}
\end{wrapfigure}
\textbf{Saraju P Mohanty} (Senior Member, IEEE) received the bachelor’s degree (Honors) in electrical engineering from the Orissa University of Agriculture and Technology, Bhubaneswar, in 1995, the master’s degree in Systems Science and Automation from the Indian Institute of Science, Bengaluru, in 1999, and the Ph.D. degree in Computer Science and Engineering from the University of South Florida, Tampa, in 2003. He is a Professor with the University of North Texas. His research is in ``Smart Electronic Systems’’ which has been funded by National Science Foundations (NSF), Semiconductor Research Corporation (SRC), U.S. Air Force, IUSSTF, and Mission Innovation. He has authored 550 research articles, 5 books, and 10 granted and pending patents. His Google Scholar h-index is 58 and i10-index is 269 with 15,000 citations. He is regarded as a visionary researcher on Smart Cities technology in which his research deals with security and energy aware, and AI/ML-integrated smart components. He introduced the Secure Digital Camera (SDC) in 2004 with built-in security features designed using Hardware Assisted Security (HAS) or Security by Design (SbD) principle. He is widely credited as the designer for the first digital watermarking chip in 2004 and first the low-power digital watermarking chip in 2006. He is a recipient of 19 best paper awards, Fulbright Specialist Award in 2021, IEEE Consumer Electronics Society Outstanding Service Award in 2020, the IEEE-CS-TCVLSI Distinguished Leadership Award in 2018, and the PROSE Award for Best Textbook in Physical Sciences and Mathematics category in 2016. He has delivered 30 keynotes and served on 15 panels at various International Conferences. He has been serving on the editorial board of several peer-reviewed international transactions/journals, including IEEE Transactions on Big Data (TBD), IEEE Transactions on Computer-Aided Design of Integrated Circuits and Systems (TCAD), IEEE Transactions on Consumer Electronics (TCE), and ACM Journal on Emerging Technologies in Computing Systems (JETC). He has been the Editor-in-Chief (EiC) of the IEEE Consumer Electronics Magazine (MCE) during 2016-2021. He served as the Chair of Technical Committee on Very Large Scale Integration (TCVLSI), IEEE Computer Society (IEEE-CS) during 2014-2018 and on the Board of Governors of the IEEE Consumer Electronics Society during 2019-2021. He serves on the steering, organizing, and program committees of several international conferences. He is the steering committee chair/vice-chair for the IEEE International Symposium on Smart Electronic Systems (IEEE-iSES), the IEEE-CS Symposium on VLSI (ISVLSI), and the OITS International Conference on Information Technology (OCIT). He has supervised 3 post-doctoral researchers, 17 Ph.D. dissertations, 28 M.S. theses, and 28 undergraduate projects.

\end{document}